\documentclass{article}
\usepackage{ijcai17}

\usepackage{times}
\usepackage{amsmath,amsfonts,amsthm}
\usepackage{graphicx}
\usepackage{caption}
\usepackage{subcaption}
\usepackage{epstopdf}
\usepackage{color}
\usepackage{float}
\usepackage{pifont}
\usepackage{algorithm}
\usepackage{algpseudocode}
\usepackage{wrapfig}
\usepackage{tabularx}
\makeatletter
\newenvironment{procedure}[1][htb]{%
    \renewcommand{\ALG@name}{Procedure}
   \begin{algorithm}[#1]%
  }{\end{algorithm}}
\makeatother

\allowdisplaybreaks

\newtheorem{theorem}{Theorem}[section]
\newtheorem{claim}[theorem]{Claim}

\newtheorem{lemma}[theorem]{Lemma}

\theoremstyle{definition}

\newtheorem{assumption}[theorem]{Assumption}

\newcommand{\EEE}{\mathbb{E}}
\newcommand{\FFF}{\mathcal{F}}

\newcommand{\eee}{\mathcal{E}}
\newcommand{\PPP}{\mathbb{P}}

\newcommand{\KKK}{\mathcal{K}}

\newcommand{\NNN}{\mathcal{N}}

\newcommand{\RRR}{\mathbb{R}}
\newcommand{\SSS}{\mathcal{S}}
\newcommand{\TTT}{\mathcal{T}}

\title{Assortment Optimization under Unknown MultiNomial Logit Choice Models}
\author{Wang Chi Cheung, David Simchi-Levi}

\begin{document}

\maketitle

\begin{abstract}
Motivated by e-commerce, we study the online assortment optimization problem. The seller offers an assortment, i.e. a subset of products, to each arriving customer, who then purchases one or no product from her offered assortment. A customer's purchase decision is governed by the underlying MultiNomial Logit (MNL) choice model. The seller aims to maximize the total revenue in a finite sales horizon, subject to resource constraints and uncertainty in the MNL choice model.  
We first propose an efficient online policy which incurs a regret $\tilde{O}(T^{2/3})$, where $T$ is the number of customers in the sales horizon. Then, we propose a UCB policy that achieves a regret $\tilde{O}(T^{1/2})$. Both regret bounds are sublinear in the number of assortments.
\end{abstract}

\section{Introduction}
Online sales are now ubiquitous in the retail industry. During an online sale, a seller offers a handpicked assortment, i.e. a subset of products, to an arriving customer. The customer's purchase decision crucially depends on her offered assortment. She first scrutinizes all the products in the assortment, then decides which product she likes the most. After that, she either purchases her favorite, or purchases nothing if her willingness-to-pay is below the price for her favorite. Such choice behavior is captured by the underlying \emph{choice model}, which has been under intense study by the economics and operations research communities \cite{Ben-AkivaL85}. 


In order to maximize the total revenue in an online sale, the seller needs to know the underlying choice model. However, the model is often not known in practice. This motivates the seller to maximize her revenue and learn the underlying choice model simultaneously. Apart from model uncertainty, the seller often faces resource constraints; when a product is sold, a certain amount of resources is consumed, and the resources cannot be replenished during the sales horizon. The seller is then forced to stop the sales process either when the sales horizon ends, or when the resources are depleted.

In this paper, we formulate a model for the online assortment optimization problem, which encompasses choice model uncertainty and resource constraints. The seller aims to minimize his \emph{regret}, i.e. the difference between the revenue earned by an oracle, who knows the underlying choice model, and the revenue earned by the seller, who is uncertain about the model. We assume an uncertain MultiNomial Logit (MNL) model, which is fundamental in the literature. We first propose an efficient policy $\text{{\sc Online}}(\tau)$ that incurs a regret $\tilde{O}(T^{2/3})$, where $T$ is the number of customers, and $\tau$ is the length of the learning phase. Then, we propose a UCB policy with a regret of $\tilde{O}(\sqrt{T})$; the UCB policy is not known to be computationally efficient. Both regret bounds are sublinear in the total number of assortments, since we exploit the special structure of MNL choice model to avoid learning all the choice probabilities assortment by assortment. 

\section{Literature Review and Our Contributions}\label{sec:lit}
$\textbf{Offline assortment optimization.}$ The MNL choice model is a fundamental model proposed by \cite{McFadden74}, and it has been the building block for many other existing choice models \cite{Ben-AkivaL85}. Assortment optimization under the MNL choice model has been actively studied. Assuming the knowledge of the underlying MNL choice model, \cite{TallurivR04} propose an efficient algorithm for computing an optimal assortment when there is no resource constraint; \cite{LiuV08} propose an efficient algorithm for computing a \emph{mixture} of assortments that achieves asymptotic optimality under resource constraints. \cite{BernsteinKX15} offer insights into the optimal assortment planning policy under resource constraints, when the product prices are equal but there are multiple types of customers.

$\textbf{Online assortment optimization.}$ Assuming uncertainty in the MNL choice model, \cite{RusmevichientongSS10} propose an online policy that incurs an instance-dependent $O(\log T)$ regret. 
\cite{SaureZ13} generalize \cite{RusmevichientongSS10} by proposing online policies with instance-dependent $O(\log T)$ regret bounds for a wider class of choice models. 
Recently, \cite{AgrawalAGZ16} provide a instance-independent regret $\tilde{O}(\sqrt{T})$ under an uncertain MNL choice model. 
However, these existing works do not incorporate resource constraints into their models, unlike ours. Our approach is based on establishing a confidence bound on the choice probability for \emph{every assortment} (cf. Lemmas \ref{lemma:goodevent}, \ref{lemma:lipschitz}), which is novel in the literature, and necessary for learning the choice model under resource constraints.

$\textbf{Budgeted bandits. }$Our online assortment optimization problem can be cast as a budgeted bandit problem, with the arm set being the allowed assortments. For budget bandit problems, \cite{Tran-ThanhCCRJ10} provide an instance-indepenedent regret bound with a resource constraint; \cite{Tran-ThanhCRJ12} and \cite{XiaDZYQ15} provide instance-dependent regret bounds for the cases of discrete and continuous resource consumption costs. \cite{XiaLQYL15} propose a Thompson Sampling based algorithm. \cite{BadanidiyuruKS13}, \cite{AgrawalD14} provide optimal instance-independent regret bounds for the problem with general resource constraints. 

A direct application of \cite{BadanidiyuruKS13} or \cite{AgrawalD14} to our problem yields a regret \emph{linear} in the number of assortments, which is often larger than the number of customers. Indeed, their policies involve testing each assortment at least once. In contrast, we exploit the special structure of the MNL choice model to achieve a regret bound sublinear in the number of assortments.

$\textbf{Combinatorial bandits. }$ Our problem can be cast as a stochastic combinatorial bandit problem with semi-bandit feedback, when we relax the resource constraints, and interpret a product and an assortment as a \emph{basic arm} and a \emph{super arm} respectively. \cite{GaiKJ12} study the combinatorial bandit problem with linear reward (i.e. a super arm's reward is the sum of its basic arms' reward), which is subsequently generalized and refined by \cite{ChenWY13} to the case with non-linear reward. The optimal regret bound is obtained by \cite{KvetonWAS14} in the case of linear reward. \cite{ChenHLLLL16} consider the generalized case when the expected reward under a super arm depends on certain random variables associated with its basic arms. \cite{XiaQMYL16} provide an instance-dependent regret bound to the combinatorial bandit problem with a resource constraint. Recent works \cite{RadlinskiKJ08}, \cite{KvetonSWA15},  \cite{KvetonWAS15} consider the problem in the cascading-feedback setting.

Apart from the presence of resource constraints (except \cite{XiaQMYL16}), our model differs from the existing combinatorial bandit literature, as our reward function is \emph{not monotonic} in the super arm and the underlying parameters. Indeed, introducing more products in an assortment does not necessarily increase the expected revenue, since the customer's attention could be diverted to less profitable products. Therefore, novel techniques are needed for achieving a sublinear regret in our setting.


\section{Problem Definition}\label{sec:def}
We formulate the online assortment optimization problem with an unknown MultiNomial Logit (MNL) choice model. The seller has a set of products $\NNN = \{1, \ldots, N\}$ for sale, and a set of resources $\KKK = \{1, \ldots, K\}$ for composing the products. The sale of one product $i$ generates a revenue of $r(i)\in [0, 1]$, but consumes $a(i, k)\in \{0, 1\}$ units of resource $k$, for each $k\in \KKK$. Product 0 is the ``no-purchase'' product; $r(0) = 0 = a(0, k)$ for all $k\in \KKK$.

The seller starts with $C(k) = Tc(k)\in \mathbb{Z}^+$ units of resource $k$ at period 1. For periods $t = 1, \ldots, T$, the following sequence of six events happen. First, a customer arrives in period $t$. Second, the seller offers an assortment $S_t \in \SSS$ to the customer, where $\mathcal{S}$ is the family of allowed assortments. Third, the seller observes that the product $I_t \in S_t \cup\{0\}$ is purchased. Fourth, the seller earns a revenue of $r(I_t)$. Fifth, the resources are consumed: for all $k\in \KKK$, $C(k) \leftarrow C(k) - a(I_t, k)$. Sixth, the seller proceeds to period $t+1$.

A customer's purchase decision is governed by the MNL choice probability function $\varphi(\cdot, \cdot | v^*)$ \cite{McFadden74}. $v^*\in \RRR^N_{>0}$ is the latent utility parameter unknown to the seller; the seller only knows that $v^*(i) \in [1/R, R]$ for all $i\in \NNN$. For $i\in S\subset \NNN$ and $v\in \RRR^N_{> 0}$, $\varphi(i, S | v)$ represents the probability of a customer purchasing $i$ when she is offered assortment $S$, and has utility parameter $v$. The probability is defined as
\begin{equation}\label{eq:choiceprobability}
\varphi(i, S|v) := \frac{v(i)}{1 + \sum_{\ell\in S}v(\ell)}. 
\end{equation}
The customer purchases nothing with the complementary probability $\varphi(0, S | v) = 1/(1 + \sum_{\ell\in S}v(\ell)) = 1 - \sum_{i\in S}\varphi(i, S | v).$ 
For $i\in\NNN\setminus S$, $S\in \SSS$, we define $\varphi(i, S| v) = 0$. The expected revenue $\sum_{i\in S}r(i)\varphi(i, S|v)$ is \emph{not} monotonic in $S$ or $v$, in contrary to the monotonicity of reward functions in the combinatorial bandit literature.

The family of allowed assortments $\SSS$ is a subfamily of $2^{\NNN}$. One common example is the cardinality constrained family $\SSS = \{S\subset \NNN: |S|\leq B\}$. We assume that $\emptyset\in \SSS$; that is, the seller can reject a customer by offering an empty assortment, for example when the resources are depleted. We denote $B = \max\{|S|:S\in\SSS\}$ as the maximum assortment size; in most setting, $B$ is much smaller than $N$, the number of products. 


$\textbf{Regret Minimization. }$The seller's objective is to design a non-anticipatory policy that maximizes the total revenue $\sum^T_{t = 1}r(I_t)$, subject to the resource constraints. This can be formulated as the minimization of regret, which is
\begin{equation}\label{eq:regret}
\text{{\sc Reg}} = T\text{{\sc Opt}}(\text{LP}(v^*)) - \sum^T_{t = 1}r\left(I_t\right),
\end{equation}
subject to the resource constraints: for all $k\in \KKK$, $\sum^T_{t = 1}a(I_t, k) \leq Tc(k)$ always. Equivalently, we require $C(k) \geq 0$ at every period. 
The purchased product $I_t$ depends on the offered assortment $S_t$ determined by the policy. We say that a policy is \emph{non-anticipatory} if the offered assortment $S_t$ depends only on the sales history as well as the seller's randomness $U_t$ in period $t$, i.e. $S_t \in \sigma(U_t, \{S_s, I_s, U_s\}^{t-1}_{s=1})$. 

For any $v\in \RRR^N_{>0}$, the linear program LP$(v)$ is defined as
\begin{subequations}\label{LP:MNL}
\begin{alignat}{3}
\text{max }    &\sum_{S\in \SSS} R(S | v )y(S) \nonumber\\
\text{s.t. }   &\sum_{S\in \SSS} A(S, k | v)y(S) \leq c(k)       &\quad &\forall k \in \KKK \nonumber\\
&\sum_{S\in \SSS}y(S) = 1, \quad y(S)\geq   0 &\quad & \forall S\in \SSS.\nonumber
\end{alignat}
\end{subequations}
We use the notation $R(S|v) = \sum_{i\in S}r(i)\varphi(i, S|v)$ to denote the expected revenue earned by offering $S$ in a period, and $A(S, k|v) = \sum_{i\in S}a(i, k)\varphi(i, S|v)$ to denote the expected amount of resource $k$ consumed in a period. The optimal value of LP$(v)$ is denoted as {\sc Opt}(LP$(v)$). By interpreting $y$ as a probability distribution over $\SSS$, LP$(v)$ is equivalent to the maximization of the expected revenue in a period, when the resource constraints hold \emph{in expectation}. LP$(v)$ is always feasible, since $y(\emptyset) = 1, y(S) = 0$ for all $S \in \SSS\setminus \{\emptyset\}$ is a feasible solution. The benchmark $T\text{{\sc Opt}}(\text{LP}(v^*))$ upper bounds the expected optimum \cite{BadanidiyuruKS13}:
\begin{theorem}[\cite{BadanidiyuruKS13}]
For any non-anticipatory policy $\pi$ that satisfies the resource constraints with probability 1, the following inequality holds:
\begin{equation*}
T\text{{\sc Opt}}(\text{LP}(v^*)) \geq \EEE\left[ \sum^T_{t = 1}r\left(I^\pi_t\right)\right].
\end{equation*}
$I^\pi_t$ denotes the random product purchased by the period $t$ customer under policy $\pi$.
\end{theorem}

\section{Online policy {\sc Online}$(\tau)$}\label{sec:policy}
We propose the non-anticipatory policy {\sc Online}$(\tau)$, where $\tau$ is the length of the learning phase. {\sc Online}$(\tau)$ enjoys the following performance guarantee:
\begin{theorem}\label{thm:simpleregret}
Suppose $\tau$ satisfies Assumption~\ref{ass:tau}. The policy {\sc Online}$(\tau)$ satisfies all resource constraints and incurs a regret at most
\begin{equation}\label{eq:regret_bound}
\tau + O\left(TRB\sqrt{\frac{N}{\tau}\log\frac{N}{\delta}}\right) + O \left(\sqrt{T\log \frac{K+1}{\delta}}\right) 
\end{equation}
with probability $1-\delta$. In particular, the choice $\tau = (TRB)^{2/3}N^{1/3}$ minimizes the regret bound up to a constant factor, yielding the bound $\tilde{O}((TRB)^{2/3}N^{1/3})$. 
\end{theorem}
Our regret bound is sublinear in $N, B$, in deep contrast with the regret bounds by applying \cite{BadanidiyuruKS13}, \cite{AgrawalD14}, which are linear in $|\SSS | = \Theta(N^B)$. For our theoretical analysis, we assume the following on $\tau$:
\begin{assumption}\label{ass:tau}
The learning phase length $\tau$ satisfies: (i) For all $k\in \KKK,$ $\tau\sqrt{\log \frac{4NK}{\delta}}\leq T c(k)$. (ii) For all $k\in \KKK,$ $C\epsilon(\tau)\leq \frac{1}{2}c(k)$, where 
\begin{equation}\label{eq:defepsilon}
\epsilon(\tau) = 4R \sqrt{\frac{N}{\tau}\log\frac{4N}{\delta}}.
\end{equation}
\end{assumption}
Assumption \ref{ass:tau} (i) ensures that no resource is depleted during the learning phase, and (ii) ensures that the learning phase is long enough for estimating $v^*$. Assumption \ref{ass:tau} is only necessary for our analysis; {\sc Online}$(\tau)$ can be implemented for any choice of $1\leq \tau \leq T$. In our simulation results in \S \ref{sec:numerical}, {\sc Online}$(T^{2/3})$ still converges to optimal, even when the assumption is violated for the choice $\tau = T^{2/3}$. (Theorem \ref{thm:simpleregret} implies a regret of $\tilde{O}(T^{2/3}R B \sqrt{N})$ \emph{if} $\tau = T^{2/3}$ satisfies Assumption \ref{ass:tau}.) We further discuss the assumption in Appendix A.



{\sc Online}$(\tau)$ is presented in Algorithm \ref{alg:MNL_Learn}. Periods $1$ to $\tau$ are the learning phase, and periods $\tau + 1$ to $T$ are the earning phase. During the learning phase, the seller offers single item assortments in order to estimate $\left\{v^*(i)\right\}_{i\in \NNN}$. When the learning phase ends, he computes the MLE $\hat{v}(i)$ for each product. $\hat{v}(i)$ is a solution to $\min_{v\in [1/R, R]} \mathcal{L}_i(v)$. The negative log likelihood $\mathcal{L}_i(v)$ is  
\begin{align}
=& -\log\left[\prod^{i\tau/N}_{s = \frac{(i-1)\tau}{N} +1} \left(\frac{v}{1 + v}\right)^{\mathsf{1}(I_s = i)}\left(\frac{1}{1+v}\right)^{\mathsf{1}(I_s = 0)}\right] \nonumber\\
= & n(i)\log\left[1+\frac{1}{v}\right] + \left(\frac{\tau}{N} - n(i)\right)\log\left[1 + v\right],\label{eq:simple-veloglikelihood}
\end{align}
where $n(i) =  \sum^{i \tau/N}_{s = ((i-1)\tau/N) + 1}  \mathsf{1}(I_s = i)$ is the number of product $i$ sold during the learning phase.

\begin{algorithm}[t]
\caption{{\sc Online}$(\tau)$}\label{alg:MNL_Learn}
\begin{algorithmic}[1]
\State Initialize $C(k) = Tc(k)$ $\forall k\in \KKK$.
\For{$i = 1, \ldots, N$} \Comment{Learning Phase}
\For{$t = (i - 1)\tau/N + 1$ to $i\tau/N$} 
\State Offer $S_t = \{i\}$, observe outcome $I_t\in \{i, 0\}$. 
\State For all $k\in\KKK$, $C(k) \leftarrow C(k) - a(I_t,k)$.
\EndFor
\State Compute the MLE $\hat{v}(i) \in \underset{v\in [1/R, R]}{\text{argmin}} \mathcal{L}_i(v)$. 
\EndFor
\State Solve LP$(\hat{v})$ for an extreme point solution $\hat{y}$. \label{alg:MNL_solvelp}
\For{$t =\tau +1, \ldots, T$}\Comment{Earning Phase}
\State Offer $S_t$ with probability $\hat{y}(S_t)$.
\State Observe outcome $I_t\in S_t\cup \{0\}$.
\State For all $k\in \KKK$, $C(k)\leftarrow C(k) - a(I_t, k)$.
\If{$\exists k\in \KKK\text{ s.t. }C(k) = 0$}
\State {\sc Abort}; offer $S = \emptyset$ till the end.
\EndIf
\EndFor
\end{algorithmic}
\end{algorithm}
After that, we solve LP$(\hat{v})$ for an extreme point solution $\hat{y}$, which can be interpreted as a probability distribution over $\SSS$. Finally, in the earning phase, we offer $S\in \SSS$ with probability $\hat{y}(S)$ each period. At the end of a period, the seller signals {\sc Abort} when some resource is depleted, i.e. $C(k)=0$. Then, the seller offers empty assortments to subsequent customers, until the end of sales horizon. This ensures that the resource constraints are satisfied with probability 1. 

$\textbf{Computational Efficiency of {\sc Online}$(\tau)$. }$The most computationally onerous step in {\sc Online$(\tau)$} is to solve LP$(\hat{v})$, which has $|\SSS| = \Theta(N^B)$ many variables. Fortunately, by \cite{LiuV08}, LP$(\hat{v})$ can be efficiently solved by the Column Generation algorithm (CG). In each iteration of CG, we solve the \emph{reduced problem} $\max_{S\in \SSS}\tilde{R}(S | \hat{v}) = \max_{S\in \SSS}\sum_{i\in S}\tilde{r}(i)\varphi(i, S | \hat{v})$, where $\tilde{r}(i)$ is a suitably defined reduced revenue coefficient for $i$. The reduced problem is polynomial time solvable for many choices of $\SSS$, such as $\SSS = \{S : |S|\leq B\}$ \cite{RusmevichientongSS10}. In our simulations in \S~\ref{sec:numerical}, CG always terminates within 50 iterations for solving LP$(\hat{v})$. Finally, the support of $\hat{y}$, which is defined as supp$(\hat{y}) := \{S\in \SSS : \hat{y}(S) > 0\}$, has size $\leq K+1$, since $\hat{y}$ is an extreme point solution to LP$(v)$. Thus, it is easy to sample $S_t$ in the earning phase. 

$\textbf{A $\tilde{O}(\sqrt{T})$ regret policy. }$ A $\tilde{O}(\sqrt{T})$ regret can be achieved by a UCB policy:
\begin{theorem}\label{thm:regret-short}
There exists a UCB policy that satisfies the resource constraints and achieves a regret of $O\left(\sqrt{T} R^3 B^{5/2} N \log{\frac{TNK}{\delta}}\right)$ with probability at least $1-\delta$. 
\end{theorem}
The design and analysis of such a UCB policy is deferred to Appendices D - G. 
Different from {\sc Online$(\tau)$}, our UCB policy is not known to be empirically efficient.

\section{Overview of the Proof for Theorem \ref{thm:simpleregret}}\label{sec:analysis}
To begin the proof, we consider the period $t_\text{last}$ of last sale. Either {\sc Abort} is signaled at the end of period $t_{\text{last}}$, or $t_{\text{last}}=T$. $t_\text{last}$ is a random variable, depending on the resource consumption in the sales horizon. Denote (\ref{eq:regret_bound}) as {\sc Bound}$(\tau)$. We analyze the regret by the following:
\begin{align}
&\PPP\left[\text{\sc Reg}\leq  \text{{\sc Bound}}(\tau)\right] \nonumber\\
\geq &\PPP\left[T \text{{\sc Opt}}(v^*) - \sum^{t_\text{stop}}_{t=1} r(I_t)\leq \text{{\sc Bound}}(\tau)\right]\nonumber\\
\overset{(*)}{\geq} &\PPP\left[T \text{{\sc Opt}}(v^*) - \sum^{T-\rho}_{t=1} r(I_t)\leq \text{{\sc Bound}}(\tau), t_\text{stop} > T-\rho\right] \nonumber\\
\overset{(\dagger)}{\geq} &\PPP\left[\underbrace{\left\{T \text{{\sc Opt}}(v^*)  - \sum^{T-\rho}_{t=\tau + 1} r(\tilde{I}_t) \leq \text{{\sc Bound}}(\tau)\right\}}_{\mathcal{E}_\text{{\sc Reg}}}\right. \nonumber\\
&\quad\left. \cap \bigcap^{K}_{k = 1}\underbrace{\left\{\tau + \sum^{T-\rho}_{t=\tau + 1}a(\tilde{I}_t, k)\leq Tc(k)\right\}}_{\mathcal{E}_k} \right] \nonumber \\
\overset{(\ddagger)}{\geq} & \PPP\left[\mathcal{E}_{\text{{\sc Reg}}} \cap \bigcap^{K}_{k=1} \mathcal{E}_{k} \mid \mathcal{E}_{\hat{v}}\right]\PPP[\mathcal{E}_{\hat{v}}] \label{eq:final_prob}.
\end{align}
To prove the Theorem, it suffices to show that the probability (\ref{eq:final_prob}) is at least $1-\delta$. 

$\textbf{Parsing the calculation above.}$ 
In step $(*)$, we consider the event $t_\text{last}\leq \rho$, where $\rho$ is the constant
\begin{equation}\label{eq:rho}
\rho = \frac{TC\epsilon(\tau)}{\min_{k\in\KKK}c(k)} + \frac{\sqrt{T\log\frac{4(K+1)}{\delta}}}{\min_{k\in\KKK}c(k)}, 
\end{equation}
and $\epsilon(\tau)$ is defined in (\ref{eq:defepsilon}). The definition of $\rho$ is motivated in the subsequent analysis. The inequality $(*)$ is evidently true, since the probability does not increase when we require the additional event $t_\text{stop} > T-\rho$ to hold. 

To ease the analysis, we decouple the revenue and the constraints at step $(\dagger)$, by considering the process  $\{\tilde{S}_t, \tilde{I}_t\}^{T-\rho}_{t=\tau + 1}$ generated in Procedure \ref{alg:tildesimple}. The samples $\tilde{S}_{\tau + 1}, \ldots, \tilde{S}_{T-\rho}$ are i.i.d., where $\PPP[\tilde{S}_t = S] = \hat{y}(S)$. The samples $\tilde{I}_{\tau + 1}, \ldots, \tilde{I}_{T-\rho}$ are independent, where $\PPP[\tilde{I}_t = i] = \varphi(i, \tilde{S}_t | v^*)$. 

The process $\{\tilde{S}_t, \tilde{I}_t\}^{T-\rho}_{t=\tau + 1}$ is closely related to the sales process $\{S_t, I_t\}^{T-\rho}_{t=\tau + 1}$ in Algorithm \ref{alg:MNL_Learn}. We remark that: (i) If $t_\text{last} > T - \rho$, $\{\tilde{S}_t, \tilde{I}_t\}^{T-\rho}_{t=\tau + 1}$ and $\{S_t, I_t\}^{T - \rho}_{t=\tau + 1}$ are identically distributed. (ii) Otherwise, when $t_\text{last}\leq T - \rho$, {\sc Abort} is signaled before or at the end of period $T - \rho$. Then, $S_t = \emptyset$, $I_t = 0$ for $t = t_\text{last} +1, \ldots, T - \rho$, which distribute differently from $\{\tilde{S}_t, \tilde{I}_t\}^{T-\rho}_{t=t_\text{last} + 1}$.

While Procedure \ref{alg:tildesimple} requires knowing $v^*$, we emphasize that these samples are only used in our analysis. In particular, Procedure \ref{alg:tildesimple} is not needed in Algorithm \ref{alg:MNL_Learn}. 

We argue that the step $(\dagger)$ is true. Now, by remark (i), $\{\tilde{S}_t, \tilde{I}_t\}^{T-\rho}_{t=\tau + 1}$ and $\{S_t, I_t\}^{T - \rho}_{t=\tau + 1}$ are identically distributed. If the event $\mathcal{E}_k$ holds for all $k$, then the amount of resource $k$ consumed by the end of period $T-\rho$ is at most $Tc(k)$ for all $k$. This means that {\sc Abort} is not yet signaled, which implies $t_\text{stop} > T-\rho$. By replacing $\{S_t, I_t\}$ with $\{\tilde{S}_t, \tilde{I}_t\}$, we can then analyze the events $\mathcal{E}_{\text{{\sc Reg}}}$, $\{\mathcal{E}_k\}^K_{k=1}$ separately, which eases our analysis.

The step $(\ddagger)$ holds, since the probability does not increase when we require the additional event $\mathcal{E}_{\hat{v}}$ to hold. The event $\mathcal{E}_{\hat{v}}$ is defined as
\begin{equation}\label{eq:goodevent}
\left\{\left|\log \frac{\hat{v}(i)}{ v^*(i)}\right| \leq \epsilon(\tau) = 4 R\sqrt{\frac{N}{\tau}\log\frac{4N}{\delta}}\text{ for all $i$.} \right\}.
\end{equation}
The event $\mathcal{E}_{\hat{v}}$ implies that MLE $\hat{v}$ is an accurate estimator for $v^*$, with the specified confidence radius. Now, we show that the probability (\ref{eq:final_prob}) is at least $1-\delta$. This is the heart of our proof for the regret bound.

\begin{procedure}[t]
\caption{Generation of $\{\tilde{S}_t, \tilde{I}_t\}^{T-\rho}_{t=\tau + 1}$}\label{alg:tildesimple}
\begin{algorithmic}[1]
\For{$t = \tau+1, \ldots, T - \rho$}
\State Sample $\tilde{S}_t\in\SSS$ according to  $\{\hat{y}(S)\}_{S\in\SSS}$.
\State Sample $\tilde{I}_t\in \tilde{S}_t\cup \{0\}$ according to $\{\varphi(i, \tilde{S}_t | v^*)\}_i$.
\EndFor
\end{algorithmic}
\end{procedure}

$\textbf{Proving that the probability (\ref{eq:final_prob})$\geq 1-\delta$.}$ This is proved by combining Lemmas \ref{lemma:goodevent}-\ref{lemma:enoughresource}. Their proofs are deferred to Appendix B. First, we argue that the MLE $\hat{v}$ is sufficiently accurate, in the sense that the event $\mathcal{E}_{\hat{v}}$ happens with high probability:
\begin{lemma}\label{lemma:goodevent}
For any $\tau \geq N$, $\PPP[\mathcal{E}_{\hat{v}}]\geq 1 - \delta/2$.
\end{lemma}
The proof involves a change of variable $v = e^\theta$, and uses the strong convexity of $\mathcal{L}_i(e^\theta)$ in $\theta$. We next bound the probability $\PPP\left[\mathcal{E}_{\text{{\sc Reg}}} \cap \bigcap^{K}_{k=1} \mathcal{E}_{k} \mid \mathcal{E}_{\hat{v}}\right]$ by the following four Lemmas. We translate the accuracy in estimating $v^*$ to the accuracy in estimating the choice probability for every assortment:
\begin{lemma}\label{lemma:lipschitz}
For all $v, v' \in \RRR^\NNN_{>0}$, $b\in [0,1]^\NNN$ and $S\subset \NNN$, the following inequality holds: $$\sum_{i \in S}b(i)\left(\varphi(i, S |v) - \varphi(i, S |v')\right) \leq \sum_{i\in S} \left|\log\frac{v(i)}{v'(i)}\right|.$$
\end{lemma}
Lemma~\ref{lemma:lipschitz} establishes the Lipschitz continuity of $\varphi(i, S|v)$ in $\log v$. Altogether, Lemmas \ref{lemma:goodevent}, \ref{lemma:lipschitz} demonstrate that the choice probability under every assortment can be learned without testing every assortment. Furthermore, the Lemmas show that $\left| R(S|\hat{v}) - R(S|v^*)\right| = O(1/\sqrt{\tau})$ for all $S\in \SSS$ 
, and that $\left| A(S, k|\hat{v}) - A(S, k|v^*)\right| = O(1/\sqrt{\tau})$ for all $S\in \SSS, k\in \KKK$. 
This leads to the following Lemma:
\begin{lemma}\label{lemma:goodupperbound}
Condition on $\mathcal{E}_{\hat{v}}$ (cf. (\ref{eq:goodevent})), we have $$
\text{{\sc Opt}}(\text{LP}(\hat{v}))  \geq \left[1 - \frac{B\epsilon(\tau)}{\min_{k\in\KKK}\{c(k)\}}\right]\text{{\sc Opt}}(\text{LP}(v^*)) - B\epsilon(\tau). $$
\end{lemma}
Assumption \ref{ass:tau} (ii) ensures that $\frac{B\epsilon(\tau)}{\min_{k\in\KKK}\{c(k)\}} < 1$. 
Using Lemma \ref{lemma:goodupperbound}, we first prove the near optimality in revenue: 
\begin{lemma}\label{lemma:goodrevenue}
We have $\PPP\left[\mathcal{E}_{\text{{\sc Reg}}}\mid \mathcal{E}_{\hat{v}}\right] \geq 1 - \frac{\delta}{2(K+1)}$.
\end{lemma}
The proof involves a decomposition of the regret  in revenue and applications of Chernoff inequality. Finally, we also argue that resource $k$ are not fully consumed before period $T - \rho$.
\begin{lemma}\label{lemma:enoughresource}
We have $\PPP\left[\mathcal{E}_k\mid \mathcal{E}_{\hat{v}}\right]\geq 1 - \frac{\delta}{2(K+1)}$ for all $k\in \KKK$.
\end{lemma}
The proof for Lemma \ref{lemma:enoughresource} is similar to the proof of Lemma \ref{lemma:goodrevenue}. Altogether, the regret bound in Theorem \ref{thm:simpleregret} is proved. 

\section{Numerical Experiments}\label{sec:numerical} 
We evaluate the performance of {\sc Online}$(T^{2/3})$ with synthetic data, with varying model parameters. By Theorem \ref{thm:simpleregret}, it incurs a regret $\tilde{O}(T^{2/3}RB\sqrt{N})$. We define a \emph{class tuple} $\Gamma$ as $(\SSS, N, K, R)$, and consider random problems  model generated based on $\{\Gamma_i\}^3_{i=1}$ and 8 sales horizon lengths $\{\TTT(q)\}^8_{q=1}$, which are defined below:
\begin{equation*}
\Gamma_1 = (\SSS_1(6), 10, 5, 3), \qquad \Gamma_2 = (\SSS_1(9),15, 6, 5),  
\end{equation*}
\begin{equation*}
\Gamma_3 = (\SSS_1(15), 25, 8, 7),
\end{equation*}
\begin{equation*}
\TTT = [250, 500, 750, 1000, 1500, 2000, 5000, 10000].
\end{equation*}
Here, we denote $\SSS_1(B) = \{S\subset \NNN: |S|\leq B\}$. The tuples $\Gamma_1, \Gamma_2, \Gamma_3$ are ordered with increasing difficulty; the number of assortments in $\Gamma_1, \Gamma_2, \Gamma_3$ are 210, 5005 and 3.27$\times 10^6$ respectively. In many cases (especially $\Gamma_3$), there are more possible assortments than the number of periods, which makes the existing budgeted bandit policies (cf. \S~\ref{sec:lit}) infeasible. 

For each $(\Gamma_i, \TTT(q))$, we generate 5 random problem models. Then, for each of the problem models, we run {\sc Online}$(\TTT(q)^{2/3})$ 200 times, over the synthetic data generated with the model. After that, for each model, we compute two quantities: (a) the average revenue-to-optimum ratio, which is the earned revenue averaged over the 200 simulation runs divided by $\TTT(q)\text{{\sc Opt}}(\text{LP}(v^*))$, and (b) the average regret, which is $\TTT(q)\text{{\sc Opt}}(\text{LP}(v^*))$ minus the earned revenue averaged over the 200 runs. Finally, for each $(\Gamma_i, \TTT(q))$, we further average the quantities (a, b) over the 5 generated models. {\sc Online}$(\tau)$ is very efficient via the use of CG (cf. \S~\ref{sec:policy}). In our simulation, CG always terminates in 50 iterations, and each run can be simulated in less than 10 seconds for models from $\Gamma_3.$
\begin{figure}[t]
\vspace{-0.2cm}
\hspace{-0.6cm}
\centering
\begin{subfigure}[b]{0.26\textwidth}
\includegraphics[width = \textwidth]{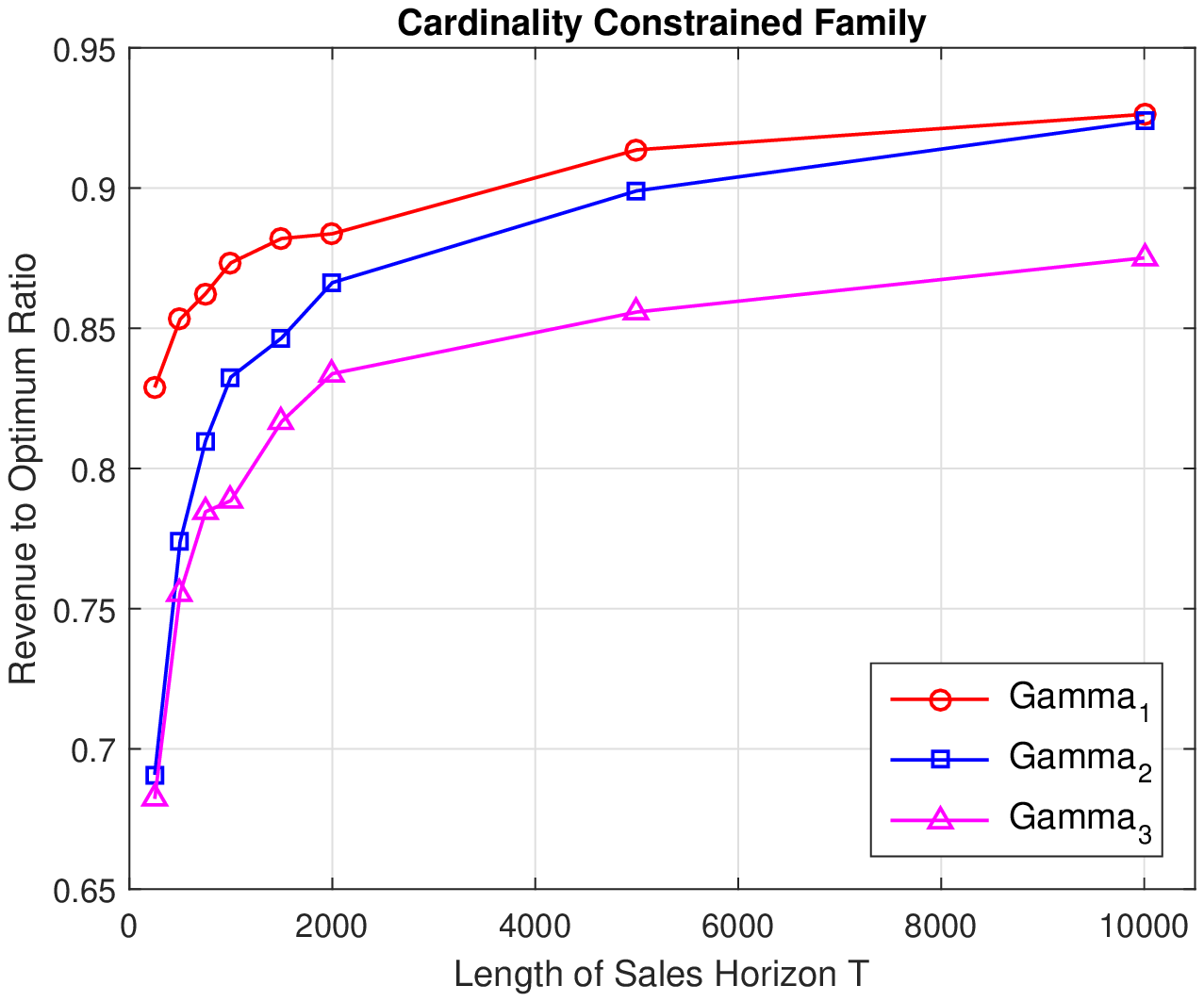}
\caption{Revenue to optimum ratios.}
\label{fig:ratio}
\end{subfigure}
\hspace{-0.6cm}
\begin{subfigure}[b]{0.26\textwidth}
\includegraphics[width = \textwidth]{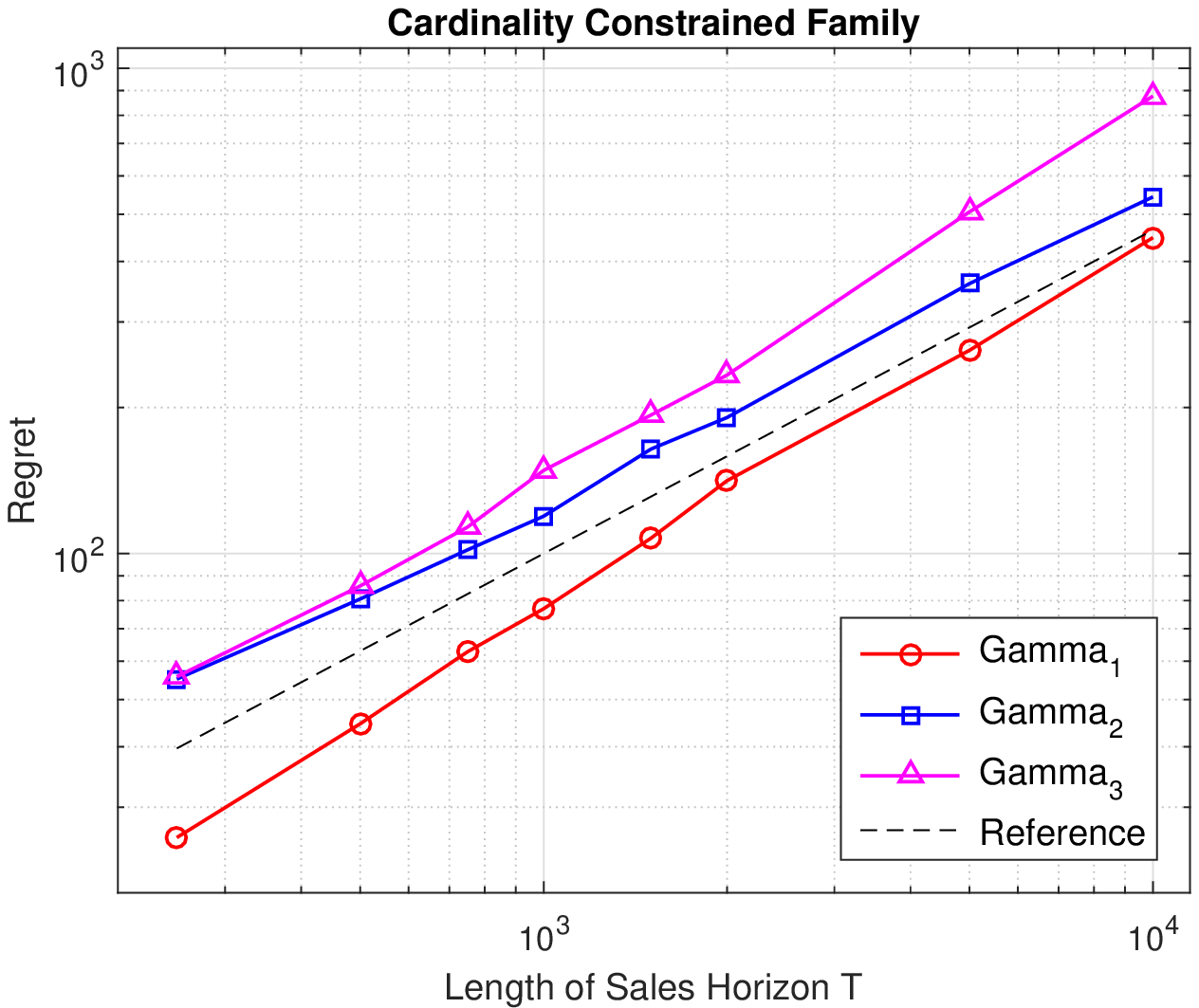}
\caption{Regret in log-log scale.}\label{fig:regret}
\end{subfigure}
\hspace{-0.6cm}
\vspace{-0.2cm}
\end{figure}

Fig.~\ref{fig:ratio} depicts the trend of the average revenue-to-optimum ratio for each $\Gamma_i$ when $ \TTT(q)$ varies. The ratio converges to 1 as $\TTT(q)$ increases. In addition, our policy performs well even when $\TTT(q)$ is small. For example, for $\Gamma_3$ where $|\SSS| = 3.27\times 10^6$, our policy is still able to achieves a ratio of 0.68 when $\TTT(q) = 250$. This demonstrates that {\sc Online}$(T^{2/3})$ performs well even when Assumption \ref{ass:tau} is violated and the sales horizon is short. After witnessing the convergence, we investigate the regret's growth rate. Fig. \ref{fig:regret} depicts the trend of the average regret for each $\Gamma_i$ when $ \TTT(q)$ varies, in the log-log scale. The black dashed line represents $f(T) = T^{2/3}$, which has slope$=2/3$. Observe that the simulated regret grows at a rate $T^{2/3}$, confirming Theorem \ref{thm:simpleregret}.

\vspace{0.1cm}
\resizebox{0.35\textwidth}{!}{
\begin{minipage}{.4\textwidth}
 \begin{tabular}{| c | c | c | c | c | c |}
    \hline
    Class $\backslash$ $T$ & 500 & 1000 & 2000 & 5000 & 10000 \\ \hline
    $(\SSS_1(4), 6, 5, 3)$ & 0.330 & 0.370 & 0.450 & 0.465	& 0.520 \\ \hline
    $(\SSS_1(6), 10, 5, 3)$ & 0.175 & 0.140 & 0.240 &  0.255 & 0.300 \\ \hline
    $(\SSS_1(9), 15, 5, 3)$ & 0.055  & 0.065  & 0.095 &  0.100 & 0.135 \\ \hline
    $(\SSS_1(10), 20, 9, 5)$ & 0.065  & 0.055  & 0.050   & 0.070   & 0.085\\ \hline
    \end{tabular}
\vspace{0.5cm}
$\text{Table 1: Fraction of instances where $\text{supp}(\hat{y}) = \text{supp}(y^*)$.}$
\end{minipage}}
\vspace{-0.3cm}

To further study the convergence of our policy, we examine the how often {\sc Online}$(T^{2/3})$ correctly identify supp$(y^*)$ after the learning phase. (Recall the notation $\text{supp}(y) = \{S\in \SSS : y(S)>0\}$.) In Table 1, for each class tuple and $T$ we tabulate the fraction of instances, out of 200 runs, where $\text{supp}(\hat{y})=\text{supp}(y^*)$. This is a stringent criterion, since supp$(\hat{y})$ could be different from supp$(y^*)$ because of multiplicity in the optimal solutions for LP$(v^*)$, and near optimality could still be achieved without $\text{supp}(\hat{y}) = \text{supp}(y^*)$. However, {\sc Online}$(T^{2/3})$ is still able to identify the support in small instances. Additional simulation results show similar trend of convergence and effectiveness in short sales horizon. The details are provided Appendix \ref{app:additionalsim}.
\section{Conclusion and Future Directions}
The online assortment optimization problem under model uncertainty and resource constraints is studied. We propose online policies, with regret bounds sublinear in the number of periods and assortments. Many interesting research directions remain to be explored. First, it is not known if the regret lower bound by \cite{AgrawalAGZ16} can be attained. Second, the incorporation of contextual information, similar to \cite{ChuLRS11}, \cite{AgrawalD16}, is an exciting topic. 

\appendix
\section{A Discussion on Assumption \ref{ass:tau}}
We remark that the choices of $\tau = T^{2/3} R^{2/3} B^{2/3} N^{1/3} $ and $\tau = T^{2/3}$ satisfy Assumption \ref{ass:tau} when $T$ is sufficiently large. Indeed, for the case of $\tau = T^{2/3} R^{2/3} B^{2/3} N^{1/3}$, Assumption \ref{ass:tau} (i, ii) are equivalent to
\begin{equation*}
T \geq \frac{ R^2 B^2}{c(k)^3}\log^{3/2}\frac{4NK}{\delta}, \quad T\geq 512\frac{ R^2 B^2 N}{c(k)^3}\log^{3/2}\frac{4N}{\delta}
\end{equation*}
for all $k\in \KKK$. For the case of $\tau = T^{2/3}$, Assumption \ref{ass:tau} (i, ii) are equivalent to
\begin{equation*}
T\geq \frac{1}{c(k)^3}\log^{3/2}\frac{4NK}{\delta}, \quad T\geq 512\frac{ B^3 R^3 N^{3/2}}{c(k)^3}\log^{3/2}\frac{4N}{\delta} 
\end{equation*}
for all $k\in \KKK$. Again for the case of $\tau = T^{2/3}$, our numerical results in \S\ref{sec:numerical} shows that {\sc Online}$(\tau)$ is effective even when the assumption is violated.
\section{Proofs for the Lemmas in Section 5}
\subsection{Proof of Lemma \ref{lemma:goodevent}}
Recall the definition of $\mathcal{L}_i(v)$ in (\ref{eq:simple-veloglikelihood}). Consider the change of variable $e^{\theta} = v$, and let $L_i(\theta) = \mathcal{L}_i(e^\theta)$. We have 
\begin{equation*}
L_i(\theta) = n(i)\log\left[1+e^{-\theta}\right] + \left(\frac{\tau}{N} - n(i)\right)\log\left[1 + e^\theta\right].
\end{equation*}
Denote $\hat{\theta} = \log \hat{v}(i)$, and $\theta^* = \log v^*(i)$.  By a Taylor Series Expansion on $f(\gamma) = L_i(\gamma \theta^* + (1-\gamma)\hat{\theta})$, we have
\begin{equation*}
L_i(\hat{\theta}) = L_i(\theta^*) + L'_i(\theta^*) (\hat{\theta} - \theta^*) + 0.5 L''_i(\breve{\theta})(\hat{\theta} - \theta^*)^2.
\end{equation*}
where $\breve{\theta} = \gamma\theta^* + (1-\gamma)\hat{\theta}$ for some $\gamma \in (0, 1)$. Since $L_i(\hat{\theta}) \leq L_i(\theta^*)$,  we have
\begin{equation}\label{eq:bettertaylor}
0\geq L'_i(\theta^*) (\hat{\theta} - \theta^*) + 0.5 L''_i(\breve{\theta})(\hat{\theta} - \theta^*)^2.
\end{equation}
Interestingly, the first derivative term can be bounded as follows:
\begin{equation}\label{eq:1stderi}
\left|L'_i(\theta^*)\right| = \left|\frac{\tau}{N}\frac{e^{\theta^*}}{1 + e^{\theta^*}} - n(i)\right| \leq \sqrt{\frac{\tau}{N}\log\frac{4N}{\delta}}
\end{equation}
with probability at least $1 - \delta/2N$. (\ref{eq:1stderi}) is by Chernoff Inequality, since $N(i)$ is a sum of $\tau/N$ i.i.d. 0-1 random variables $\{\mathsf{1}(I_s = i)\}^{\tau/N}_{s=1}$, which has expectation $e^{\theta^*}/(1 + e^{\theta^*})$. Next, we bound the second derivative as follows:
\begin{equation}\label{eq:2ndderi}
L''_i(\breve{\theta}) = \frac{\tau e^{\breve{\theta}(i)}}{N(1 + e^{\breve{\theta}(i)})^2}\geq \frac{R}{N (1 + R)^2}.
\end{equation}
Combining (\ref{eq:bettertaylor}, \ref{eq:1stderi}, \ref{eq:2ndderi}) and substituting $v^*(i), \hat{v}(i)$, we have
\begin{equation*}
\frac{\tau R}{2N(1 + R)^2}\left|\log\frac{\hat{v}(i)}{v^*(i)}\right|^2 - \sqrt{\frac{\tau}{N}\log\frac{4N}{\delta}}\left|\log\frac{\hat{v}(i)}{v^*(i)}\right| \leq 0.
\end{equation*}
with probability at least $1 - \delta/2N$. Finally, the Lemma is proved by taking a union bound over all products.

\subsection{Proof of Lemma \ref{lemma:lipschitz}}
Consider function $f:[0, 1]\rightarrow \RRR$ defined by $f(\gamma) = \sum_{i \in S}b(i)\left(\varphi(i, S |\exp\left[\theta' + \gamma(\theta - \theta')\right]\right)$, where $\theta' = (\theta'(i))_{i\in \NNN} = (\log [v'(i)])_{i\in \NNN}$, and $\theta = (\theta(i))_{i\in \NNN} = (\log [v(i)])_{i\in \NNN}$. Let's also define the shorthand $\theta_\gamma(i) = \theta'(i) + \gamma(\theta(i) - \theta'(i))$. Note that $\theta_0 = \theta'$ and $\theta_1 = \theta$. By the mean value theorem, 
\begin{align}
&\sum_{i \in S}b(i)\left(\varphi(i, S |v) - \varphi(i, S |v')\right) \nonumber\\
=&f(1) - f(0) =f'(\gamma) \quad\text{for some $\gamma\in(0, 1)$}\nonumber\\
=&\sum_{i\in S}\frac{b(i)e^{\theta_\gamma(i)}}{1+ \sum_{\ell\in S}e^{\theta_\gamma(\ell)}}(\theta(i) - \theta'(i)) \nonumber\\
\quad &- \sum_{i\in S}e^{\theta_\gamma(i)}\frac{\sum_{j\in S}b(j)e^{\theta_\gamma(j)}}{(1+ \sum_{\ell\in S}e^{\theta_\gamma(\ell)})^2}(\theta(i) - \theta'(i))\label{eq:stepdifference}\\
\leq & \sum_{i\in S} \left|\theta(i) - \theta'(i)\right| = \sum_{i\in S} \left|\log\frac{v(i)}{v'(i)}\right|.\nonumber
\end{align}
The inequality (\ref{eq:stepdifference}) holds since the sum of the coefficients of $(\theta(i) - \theta'(i))$ in the two summations lies in $[0, 1]$.

\subsection{Proof of Lemma \ref{lemma:goodupperbound}} \label{app:pflemmagoodupperbound}
Consider the following linear program S-LP:
\begin{subequations}
\begin{alignat}{3}
\text{max }    &\sum_{S\in \SSS} \left[R(S|\hat{v}) + B\epsilon(\tau)\right]y(S) \label{eq:S-LP.1}\\
\text{s.t. }   &\sum_{S\in \SSS} \left[A(S, k | \hat{v}) - B\epsilon(\tau)\right]y(S) \leq c(k) -  B\epsilon(\tau)   &\quad &\forall k \in \KKK \label{eq:S-LP.2}\\
&\sum_{S\in \SSS}y(S) = 1, \quad y(S)\geq   0 &\quad & \forall S\in \SSS \label{eq:S-LP.3},
\end{alignat}
\end{subequations}
and let {\sc Opt}(S-LP) denote its optimal value. We claim the following, conditional on the event $\mathcal{E}_{\hat{v}}$:
\begin{align}
&\text{{\sc Opt}}(LP(\hat{v}))+ B\epsilon(\tau)\nonumber\\
= &\text{{\sc Opt}(S-LP)} \label{eq:S-LPstep1}\\
\geq &\left(1 - \frac{B\epsilon(\tau)}{\min_{k\in \KKK}\left\{c(k)\right\}}\right)\text{{\sc Opt}}(LP(v^*)).\label{eq:S-LPstep2}
\end{align}

$\textbf{Proving (\ref{eq:S-LPstep1}): }$Rearranging the constraint for resource $k$ yields $\sum_{S\in \SSS}A(S, k|\hat{v})y(S)\leq c(k)$, which is the resource $k$ constraint for LP$(\hat{v})$. Similarly, the objective of S-LP is equal to the objective of LP$(\hat{v})$ plus $B\epsilon(\tau)$. This proves (\ref{eq:S-LPstep1}).

$\textbf{Proving (\ref{eq:S-LPstep2}): }$ Define the shorthand $\kappa = \frac{B\epsilon(\tau)}{\min_{k\in \KKK}\{c(k)\}}$. We first claim that the solution 
\[\breve{y}(S) = \left\{
  \begin{array}{lr}
    \left(1 - \kappa\right)y^*(S) &\text{ if } S\in \SSS\setminus \emptyset\\
     \kappa + \left(1 - \kappa\right)y^*(\emptyset)&\text{ if } S= \emptyset
  \end{array}
\right.
\]
is feasible to S-LP, where $y^*$ is an optimal solution to LP$(v^*)$. Given the feasibility of $\breve{y}$ to S-LP, we have
\begin{align}
\text{{\sc Opt}(S-LP)}&\geq \sum_{S\in \SSS} \left[R(S|\hat{v}) + B\epsilon(\tau)\right]\breve{y}(S)\nonumber\\
&\geq \sum_{S\in \SSS} R(S|v^*)\breve{y}(S) \label{eq:appKstep0}\\
&= \left(1 - \frac{B\epsilon(\tau)}{\min_{k\in \KKK}\{c(k)\}}\right)\sum_{S\in \SSS} R(S|v^*)y^*(S) \nonumber\\
&= \left(1 - \frac{B\epsilon(\tau)}{\min_{k\in \KKK}\{c(k)\}}\right)\text{{\sc Opt}}(LP(v^*))\nonumber.
\end{align}
Step (\ref{eq:appKstep0}) is justified as follows. Conditional the event $\mathcal{E}_{\hat{v}}$, Lemma \ref{lemma:lipschitz} implies that, for all $S\in \SSS$ we have 
\begin{equation}\label{eq:appKlipschitz}
\left|R(S|\hat{v}) - R(S|v^*)\right|\leq B\epsilon(\tau).
\end{equation}
This justifies the step (\ref{eq:appKstep0}).

Finally, we return to checking the feasibility $\breve{y}$. First, the constraints in (\ref{eq:S-LP.3}) hold; in particular, the equality $\sum_{S\in \SSS}\breve{y}(S) = 1$ holds by our definition of $\breve{y}(\emptyset)$. Note that the factor $\left(1 - \frac{B\epsilon(\tau)}{\min_{k\in \KKK}\left\{c(k)\right\}}\right)$ is non-negative, by Assumption \ref{ass:tau} (ii).

To check the constraints in (\ref{eq:S-LP.2}), we have
\begin{align}
&\sum_{S\in \SSS} \left[A(S|\hat{v}) -B\epsilon(\tau)\right]\breve{y}(S)  \nonumber\\
= &\left(1 - \frac{B\epsilon(\tau)}{\min_{k\in \KKK}\left\{c(k)\right\}}\right)\sum_{S\in \SSS} \left[A(S, k | \hat{v}) - B\epsilon(\tau)\right]y^*(S)\nonumber\\
\leq &\left(1 - \frac{B\epsilon(\tau)}{\min_{k\in \KKK}\left\{c(k)\right\}}\right)\sum_{S\in \SSS} A(S, k|v^*)y^*(S)\label{eq:appKstep1}\\
\leq &\left(1 - \frac{B\epsilon(\tau)}{\min_{k\in \KKK}\left\{c(k)\right\}}\right)c(k) \label{eq:appKstep2}\\
\leq & c(k) - B\epsilon(\tau)\nonumber,
\end{align}
where (\ref{eq:appKstep1}) is by (\ref{eq:appKlipschitz}), and (\ref{eq:appKstep2}) is by the feasibility of $y^*$ to LP$(v^*)$. Altogether, $\breve{y}$ is feasible to S-LP, and this finishes the proof of the Lemma.

\subsection{Proof of Lemma \ref{lemma:goodrevenue}}
Recall the shorthand $\kappa = B\epsilon(\tau)/\min\{c(k)\}$ used in Appendix B.3. Conditional on $\mathcal{E}_{\hat{v}}$, we have:
\begin{align}
&T \text{{\sc Opt}}(LP(v^*)) - \sum^{T-\rho}_{t=\tau +1} r(\tilde{I}_t) \nonumber\\
\leq& T\left(\text{{\sc Opt}}(LP(\hat{v})) + \kappa\text{{\sc Opt}}(LP(v^*)) + B\epsilon(\tau)\right) -\sum^{T-\rho}_{t=\tau + 1}r(\tilde{I}_t)\label{eq:appKbyL}\\
\leq &\rho + \tau + (T-\rho-\tau)\left(\kappa\text{{\sc Opt}}(LP(v^*)) + B\epsilon(\tau)\right) \nonumber\\
&\quad + \underbrace{(T- \rho-\tau)\text{{\sc Opt}}(LP(\hat{v})) - \sum^{T-\rho}_{t=\tau + 1}r(\tilde{I}_t)}_{\text{{\sc (Regret)}}} \label{eq:appKregret}.
\end{align}
The inequality (\ref{eq:appKbyL}) is by Lemma \ref{lemma:goodupperbound}. We decompose the term ({\sc Regret}) as follows:
\begin{align}
&(\text{{\sc Regret}})= \underbrace{(T- \rho-\tau)\text{{\sc Opt}}(LP(\hat{v}))- \sum^{T-\rho}_{t = \tau+ 1}R(\tilde{S}_t|\hat{v})}_{(\heartsuit_0)}\nonumber\\
\quad & +\underbrace{ \sum^{T-\rho}_{t = \tau+ 1} R(\tilde{S}_t|\hat{v}) -  \sum^{T-\rho}_{t = \tau+ 1}R(\tilde{S}_t|v^*)}_{(\clubsuit_0)} \nonumber\\
&\qquad +\underbrace{\sum^{T-\rho}_{t = \tau+ 1}R(\tilde{S}_t|v^*) - \sum^{T-\rho}_{t = \tau+ 1}r(\tilde{I}_t)}_{(\diamondsuit_0)}.\nonumber 
\end{align}
We prove the following the bounds for $(\heartsuit_0, \clubsuit_0, \diamondsuit_0)$.

$\textbf{To bound $(\heartsuit_0)$:}$ Recall {\sc Opt}$(\text{LP}(\hat{v})) = \sum_{S\in\SSS} R(S|\hat{v})\hat{y}(S)$. By the definition of the sampling procedure, we have $\PPP[\tilde{S}_t = S] = \hat{y}(S)$. For any fixed $\hat{v}$ the random variables $$\left\{ \sum_{S \in \SSS} R(S|\hat{v})\hat{y}(S) - R(\tilde{S}_t|\hat{v})\right\}^{T-\rho}_{t = \tau+1}$$ are i.i.d., mean 0, and lie in the interval $[-1, 1]$. By Chernoff bound, we have
\begin{equation*}
\PPP\left[(\heartsuit_0) \leq \sqrt{2T\log\frac{4(K+1)}{\delta}}\mid \mathcal{E}_{\hat{v}}\right] \geq 1 - \frac{\delta}{4(K+1)}.
\end{equation*}
$\textbf{To bound $(\clubsuit_0)$:}$ Recall the assumption that $r(i)\in [0,1]$, and $|\tilde{S}_t|\leq B$ for all $t$. Conditional on $\mathcal{E}_{\hat{v}}$, we have $\left|\log\frac{\hat{v}(i)}{v^*(i)}\right|\leq \epsilon(\tau)$ for every product $i$. By applying Lemma \ref{lemma:lipschitz}, this implies that for all $S$, we have 
$$R(S | \hat{v}) - R(S | v^*)\leq B\epsilon(\tau).$$
Thus, we have $(\clubsuit_0)\leq TB\epsilon(\tau).$
 
$\textbf{To bound $(\diamondsuit_0)$:}$ For any realized samples $\{\tilde{S}_t\}^{T-\rho}_{t = \tau  +1}$, the sequence $\{R(\tilde{S}_t|v^*) - r(\tilde{I}_t)\}^{T-\rho}_{t = \tau + 1}$ of random variables are independent, by the way $\tilde{I}_t$ are sampled in Procedure \ref{alg:tildesimple}. Moreover, the random variable $R(\tilde{S}_t|v^*) - r(\tilde{I}_t)$ has mean zero, and lies in the range $[-1, 1]$. By Chernoff inequality, we have $$\PPP\left[(\diamondsuit_0)\leq \sqrt{2T\log{\frac{4(K+1)}{\delta}}}\mid \mathcal{E}_{\hat{v}}\right] \geq 1 - \frac{\delta}{4(K+1)} .$$

Finally, we derived the desired bound in the Lemma. Conditional on $\mathcal{E}_{\hat{v}}$, we have
\begin{align}
&(\text{{\sc Regret}})\nonumber\\
= &T\text{{\sc Opt}}(LP(v^*)) - \sum^{T-\rho}_{t = r+1}r(\tilde{I_t})\nonumber\\
\leq &\rho + \tau + (T-\rho-\tau)B\epsilon(\tau)\left(\frac{\text{{\sc Opt}}(LP(v^*))}{\min_{k\in\KKK}\{c(k)\}} + 1\right) \nonumber\\
&\quad + TB\epsilon(\tau) + 2\sqrt{2T\log\frac{4(K+1)}{\delta}}\nonumber\\
\leq &\rho + \tau + TB\epsilon(\tau)\left(\frac{1}{\min_{k\in\KKK}\{c(k)\}} + 2\right)  \nonumber\\
&\quad + 2\sqrt{2T\log\frac{4(K+1)}{\delta}}\nonumber\\
=&\tau + TB\epsilon(\tau)\left(\frac{2}{\min_{k\in\KKK}\{c(k)\}} + 2\right) \nonumber\\
&\quad +\left(2 + \frac{1}{\min_{k\in \KKK}c(k)}\right)\sqrt{2T\log\frac{4(K+1)}{\delta}} \label{eq:appL2}
\end{align}
holds with probability $1-\delta/2(K+1)$. The step (\ref{eq:appL2}) is by the definition of $\rho$ in (\ref{eq:rho}). By the definition of $\epsilon(\tau)$, the Lemma is proved. 


\subsection{Proof of Lemma \ref{lemma:enoughresource}}\label{app:pflemmaenoughresource}
Similar to the proof for Lemma \ref{lemma:goodrevenue}, we decompose the sum $\sum^{T - \rho}_{t = \tau + 1} a(\tilde{I}_t, k)$ into 4 terms, $(\diamondsuit_k), (\clubsuit_k), (\heartsuit_k), (\spadesuit_k)$:
\begin{align}
& \sum^{T-\rho}_{\tau+1}a(\tilde{I}_t, k) = \underbrace{\sum^{T-\rho}_{t=\tau+1}a(\tilde{I}_t, k) - A(\tilde{S}_t, k| v^*)}_{(\diamondsuit_k)} \nonumber\\
\quad & +\underbrace{\sum^{T-\rho}_{t = \tau+1} A(\tilde{S}_t, k | v^*) - A(\tilde{S}_t, k | \hat{v})}_{(\clubsuit_k)} \nonumber\\
\quad & \quad + \underbrace{\sum^{T-\rho}_{t = \rho+1} A(\tilde{S}_t,k | \hat{v}) - \sum^{T-\rho}_{t = \tau +1}\sum_{S\in\SSS}A(S, k | \hat{v})\hat{y}(S)}_{(\heartsuit_k)} \nonumber\\
\quad & \qquad + \underbrace{\sum^{T-\rho}_{t = \tau +1}\sum_{S\in\SSS}A(S, k | \hat{v})\hat{y}(S)}_{(\spadesuit_k)} \nonumber 
\end{align}

We bound each term from above, conditional on $\mathcal{E}_{\hat{\theta}}$, as follows:

$\textbf{To bound $(\diamondsuit_k)$:}$ For any fixed sequence of assortments $\{\tilde{S}_t\}^{T-\rho}_{t = \tau + 1}$, the random variables $\{a(\tilde{I}_t, k) - A(\tilde{S}_t, k | v^*)\}^{T-\rho}_{t = \tau + 1}$, where $\tilde{I}_t\sim \tilde{S}_t$ are independent. Each of the random variables $a(\tilde{I}_t, k) - A(\tilde{S}_t, k | v^*)$ has mean zero, and lies in the range $[-1, 1]$. By Chernoff Bound, for any $\{\tilde{S}_t\}^{T-\rho}_{t = \tau + 1}$ the following inequality holds with probability $1 - \frac{\delta}{4(K+1)}$: 
$$\sum^{T-\rho}_{t=\tau+1}a(\tilde{I}_t, k) - a(\tilde{S}_t, k|v^*) \leq \sqrt{2T\log{\frac{4(K+1)}{\delta}}}. $$
In particular, this is true condition on $\mathcal{E}_{\hat{v}}$, hence proving that $$\PPP\left[(\diamondsuit_k)\leq \sqrt{2T\log{\frac{4(K+1)}{\delta}}}\mid \mathcal{E}_{\hat{v}}\right] \geq 1 - \frac{\delta}{4(K+1)} .$$

$\textbf{To bound $(\clubsuit_k)$:}$ We bound $(\clubsuit_k)$ in a similar way to the case of $(\clubsuit_0)$. Now, $a(i, k)\in \{0,1\}$, and $|\tilde{S}_t|\leq B$ for all $t$. Conditional on $\mathcal{E}_{\hat{v}}$, we have $\left|\log\frac{\hat{v}(i)}{v^*(i)}\right|\leq \epsilon(\tau)$. By Lemma \ref{lemma:lipschitz}, for all $i, S$ we have 
$$A( S, k | v^*) - A(S, k | \hat{v})\leq B\epsilon(\tau).$$
Thus, we have $(\clubsuit_k)\leq TB\epsilon(\tau).$
 
$\textbf{To bound $(\heartsuit_k)$:}$ Recall that $\PPP[\tilde{S}_t = S] = \hat{y}(S)$ (cf. Procedure \ref{alg:tildesimple}). For any fixed $\hat{v}$ the random variables $$\left\{A(\tilde{S}_t, k|\hat{v}) - \sum_{S\in\SSS}A(S, k|\hat{v})\hat{y}(S)\right\}^{T-\rho}_{t = \rho+1}$$ are i.i.d., mean 0, and lie in the interval $[-1, 1]$. By Chernoff bound, we have
\begin{equation*}
\PPP\left[(\heartsuit_k) \leq \sqrt{2T\log\frac{4(K+1)}{\delta}}\mid \mathcal{E}_{\hat{v}}\right] \geq 1 - \frac{\delta}{4(K+1)}.
\end{equation*}

$\textbf{To bound $(\spadesuit_k)$:}$ Recall that $(\spadesuit_k)\leq (T-\rho-\tau)c(k)$, since $\hat{y}$ is a feasible solution to LP$(\hat{v})$.

Altogether, conditional on $\mathcal{E}_{\hat{v}}$, the following holds with probability $1-\delta/2(K+1)$:
\begin{align}
& (\heartsuit_k) + (\clubsuit_k) + (\diamondsuit_k) + (\spadesuit_k)\nonumber\\
\leq & 2\sqrt{2T\log{\frac{K+1}{\delta}}} + TB\epsilon(\tau) + (T-\rho-\tau)c(k)\nonumber\\
\leq & Tc(k) - \tau \label{eq:bydefofrho}
\end{align}
where (\ref{eq:bydefofrho}) is by the definition of $\rho$ (cf. (\ref{eq:rho})). Altogether, the Lemma is proved.

\section{Additional Simulation Results}\label{app:additionalsim}
We evaluate the performance of {\sc Online}$(T^{2/3})$ with synthetic data, when the family of allowable assortments is a partition matroid. Recall that a class tuple is $(\SSS, N, K, R)$. Define the notation $\SSS_2(p , b) = \{S\subset \NNN: |S\cap \NNN_j |\leq b \text{ for all }1\leq j\leq p\}$, which denotes a partition matroid assortment family. Here, $\{\NNN_1, \ldots \NNN_p\}$ is a partition of $\NNN$ into $p$ equal size subsets, where $\NNN_j = \{(N(j-1)/p) + 1, \ldots, Nj/p\}$. (Thus, we implicitly assume that $N$ is divisible by $p$). By \cite{DavisGT13}, the optimization problem $\max_{S\in \SSS_2(p, b)}R(S | v)$ is polynomial time solvable, for any  $v, p, b$. Therefore, CG can still be efficiently implemented for {\sc Online}$(T^{2/3})$. (cf the discussion on the computational efficiency CG in Section \ref{sec:policy}).

We consider random models generated according to the following class tuples:
\begin{align*}
&\Gamma_4 = (\SSS_2(2, 3), 10, 5, 3), \quad \Gamma_5 = (\SSS(3, 3), 15, 6, 5), \\
&\qquad \qquad \quad \Gamma_6 = (\SSS_2(5, 3), 25, 8, 7).
\end{align*}
Similar to Section \ref{sec:numerical}, we evaluate the performance of {\sc Online}$(\tau)$ on the problem instances with the following lengths of sales horizon:
\begin{equation*}
\TTT = [250, 500, 750, 1000, 1500, 2000, 5000, 10000].
\end{equation*}
Our evaluation procedure is completely identical to the procedure in Section \ref{sec:numerical}. Figure 2 and Table 2 have the same interpretation as Figure 1 and Table 1. Evidently, the simulation performance for partition matroid assortment families is similar to the performance for cardinality constrained assortment families.

\begin{figure}[t]
\vspace{-0.2cm}
\hspace{-0.6cm}
\centering
\begin{subfigure}[b]{0.26\textwidth}
\includegraphics[width = \textwidth]{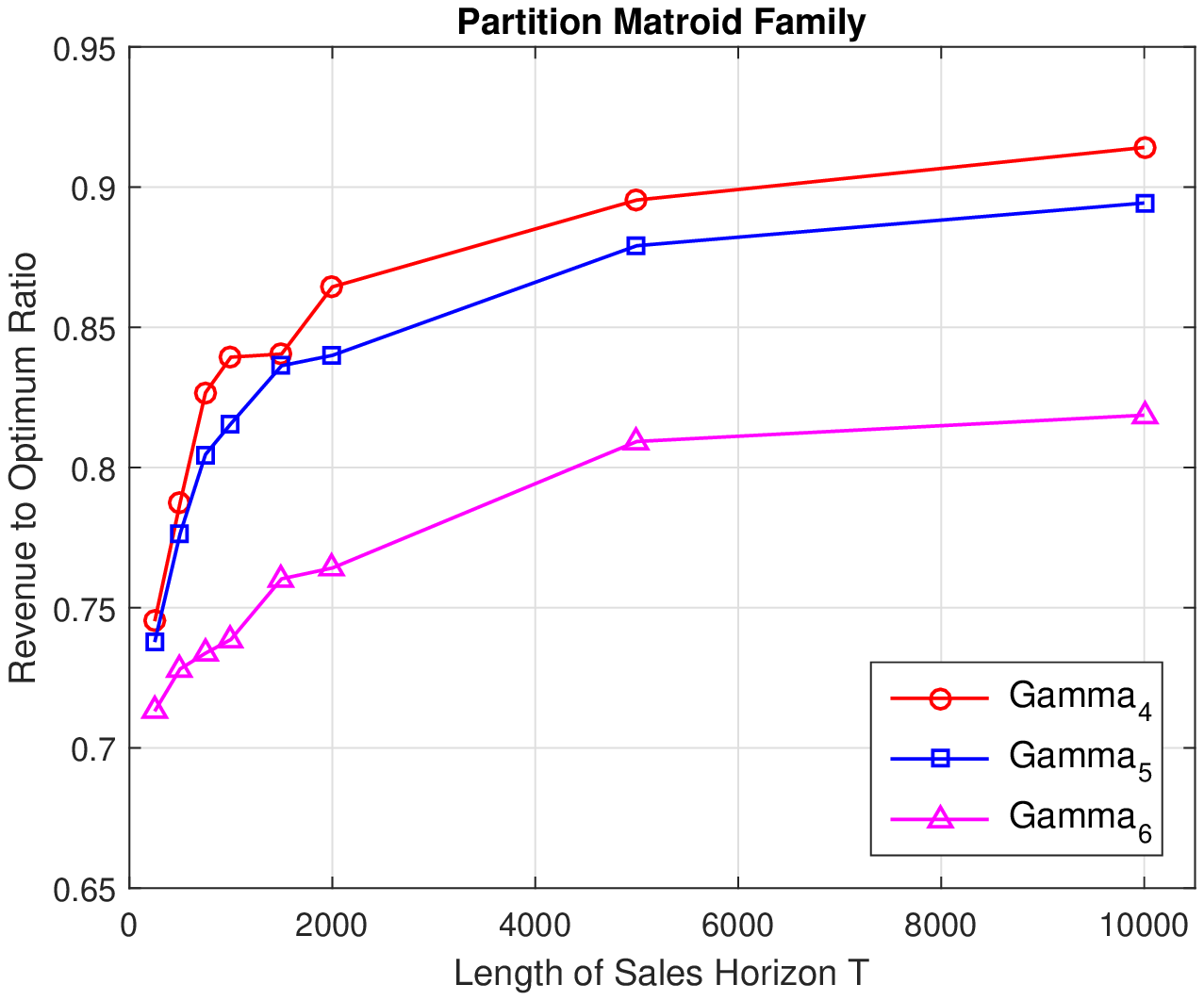}
\caption{Revenue to optimum ratios.}
\label{fig:part-ratio}
\end{subfigure}
\hspace{-0.6cm}
\begin{subfigure}[b]{0.26\textwidth}
\includegraphics[width = \textwidth]{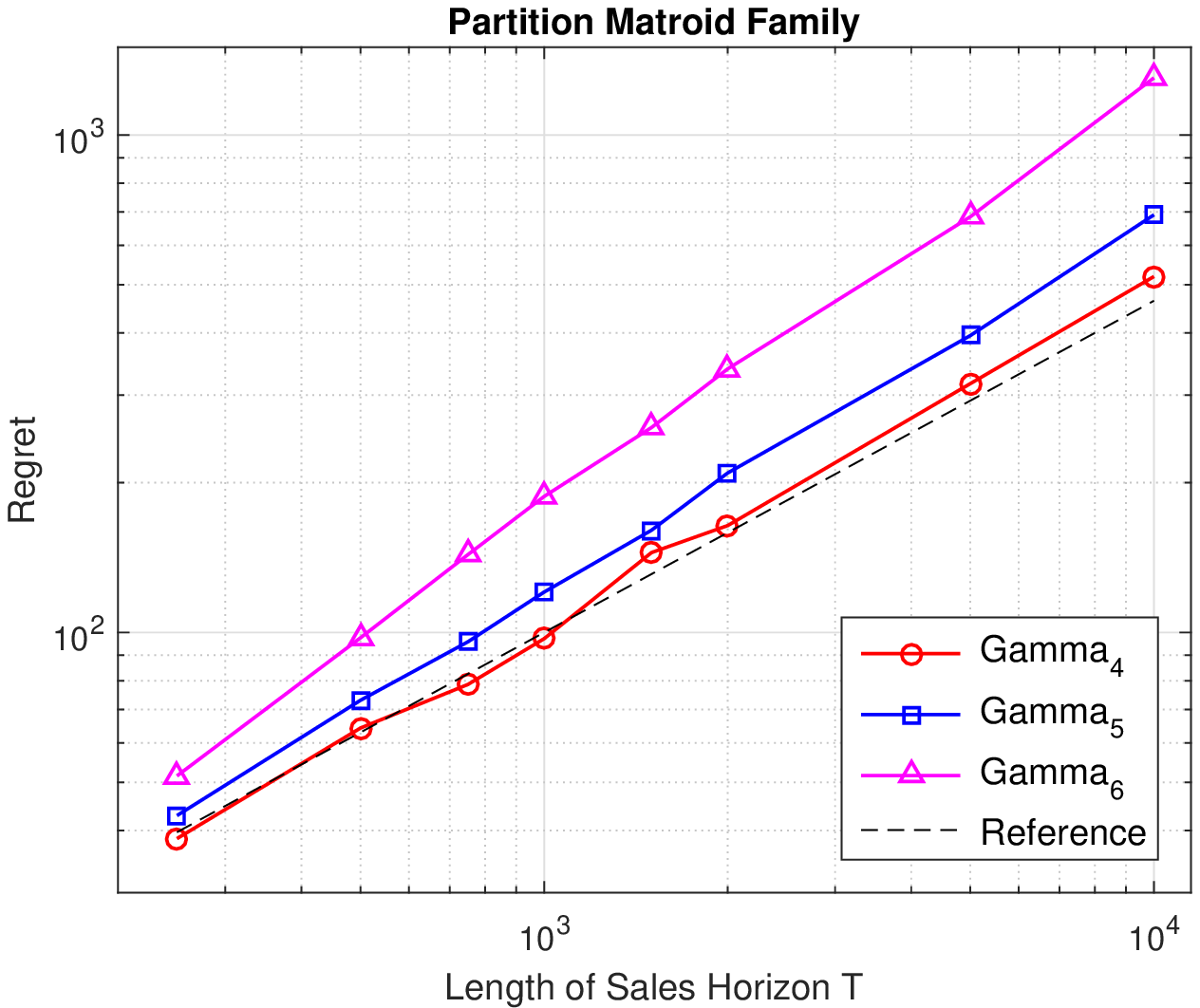}
\caption{Regret in log-log scale.}\label{fig:part-regret}
\end{subfigure}
\hspace{-0.6cm}
\vspace{-0.2cm}
\end{figure}

\vspace{0.1cm}
\resizebox{0.35\textwidth}{!}{
\begin{minipage}{.4\textwidth}
 \begin{tabular}{| c | c | c | c | c | c |}
    \hline
    Class $\backslash$ $T$ & 500 & 1000 & 2000 & 5000 & 10000 \\ \hline
    $(\SSS_2(2,2), 6, 5, 3)$ & 0.445 &	0.500 &	0.505 & 0.600 &	0.610\\ \hline
    $(\SSS_2(2,2), 10, 5, 3)$ & 0.230 & 0.300 &	0.355 & 0.385 & 0.450 \\ \hline
    $(\SSS_2(3,3), 15, 5, 3)$ & 0.140 &	0.115 & 0.190 &	0.230 &	0.325 \\ \hline
    $(\SSS_2(4,3), 20, 9, 5)$ & 0.030 &	0.040 &	0.050 &	0.075 &	0.105 \\ \hline
    \end{tabular}
\vspace{0.5cm}
$\text{Table 2: Fraction of instances where $\text{supp}(\hat{y}) = \text{supp}(y^*)$.}$
\end{minipage}}
\vspace{-0.3cm}

\section{An Online Algorithm with $\tilde{O}(\sqrt{T})$ regret}\label{sec:optimism}
In this Appendix Section, we propose and analyze a UCB policy that achieves a $\tilde{O}(\sqrt{T})$ regret. The following statement is the full version of Theorem \ref{thm:regret-short}:
\begin{theorem}\label{thm:regret}
Assume that $\omega< 1$, where $\omega$ is defined in (\ref{eq:omega}). Algorithm \ref{alg:UCB} satisfies the resource constraints with probability 1, and achieves a regret of 
$$O\left(\sqrt{T} R^3 B^{5/2} N  \log{\frac{TNK}{\delta}}\right)$$
with probability at least $1-\delta$. 
\end{theorem}
The assumption of $\omega < 1$ ensures that the sales horizon is long enough for sufficient learning. We further explain the rationale behind the assumption in the analysis.

We remark that, while the regret bound for the UCB policy (Algorithm \ref{alg:UCB}) has a better dependence on $T$ than {\sc Online}$(\tau)$, the former has a poorer dependence on $R, B, N$ than the latter. It is because the UCB policy estimates the underlying utility parameter $v^*$ with a stream of assortments $S_1, S_2, \ldots$ (and the corresponding purchase outcomes $I_1, I_2, \ldots$) of arbitrary sizes. Thus, the UCB policy needs to disentangle the dependence between different products in each offered assortment during the estimation. This situation is in contrast to {\sc Online}$(\tau)$, which estimates $v^*$ by inferring from single item assortments (and the corresponding purchase outcome). Therefore, {\sc Online}$(\tau)$ does not need to go through the disentangling process, leading to a better dependence on the parameters $R, B, N$ than our UCB policy. Nevertheless, since {\sc Online}$(\tau)$ separates learning from earning, its dependence on $T$ is strictly worse than the UCB policy, which simultaneously learns and earns.


Our UCB policy is stated in Algorithm \ref{alg:UCB}. The signal {\sc Abort} ensures that the resource constraints are satisfied with probability 1. 

\begin{algorithm}[H]
\caption{UCB Policy}\label{alg:UCB}
\begin{algorithmic}[1]
\State Initialize $C(k) = Tc(k)$ $\forall k\in \KKK$, and fixed assortments $\{S_i\}^N_{i=1}$ such that $i\in S_i\in \SSS$. \label{alg:UCB_line1}
\For{$t = 1, \ldots, N$}\Comment{Warm Start}
\State Offer $S_t$, and observe $I_t$. 
\State For all $k\in \KKK$, $C(k)\leftarrow C(k) - a(I_t, k)$. \label{alg:UCB_resource}
\EndFor 
\For{$t = N+1, \ldots, T$}\label{alg:UCB_secondfor}
\State Compute the MLE $v_t$ in (\ref{eq:min-velld}), based on $\{(S_s, I_s)\}^{t-1}_{s = 1}$.
\State Solve UCB-LP$(v_t, n_{t-1}, \omega)$ for an optimal  $\hat{y}_t$.
\State Offer an assortment $S_t\in \SSS$ with probability $\hat{y}_t(S_t)$.
\State Observe the product $I_t$ purchased.
\State For all $k\in \KKK$, $C(k)\leftarrow C(k) - a(I_t, k)$. \label{alg:UCB_resource2}
\If{$\exists k\in \KKK\text{ s.t. }C(k) = 0$}
\State Signal {\sc Abort}, break the \textbf{for}-loop and offer $S = \emptyset$ to the remaining customers.
\EndIf
\EndFor
\end{algorithmic}
\end{algorithm}
In the first $N$ periods, we warm-start our estimation on $v^*$ by offering assortments containing each of the products. Then, in each of the periods $N+1, \ldots, T$, we compute the MLE $v_t$ for $v^*$ using the observed sales history $\{(S_s, I_s)\}^{t-1}_{s = 1}$:
\begin{equation}\label{eq:min-velld}
v_t = \underset{ v \in[1/R, R]^N }{\text{argmin}}\mathcal{L}_{t - 1}(v) , 
\end{equation}
where 
\begin{equation}\label{eq:-veloglikelihood}
\mathcal{L}_{t-1}(v) = \sum^{t-1}_{s=1} - \log(\varphi(I_s, S_s|v)).
\end{equation}
After that, we solve the following UCB-LP$(v_t, n_{t-1}, \omega)$ 
\begin{subequations}\label{LP:MNL_UCB}
\begin{alignat}{3}
\text{max }    &\sum_{S\in \SSS} \left(R(S\mid v_t) + \sum_{i\in S} \varepsilon(n_{t-1}(i))\right)y(S) \nonumber\\
\text{s.t. }   &\sum_{S\in \SSS} \left(A( S\mid v_t) - \sum_{i\in S} \varepsilon(n_{t-1}(i))\right)y(S) & & \nonumber\\
& \qquad \leq (1-\omega)c(k)       &\quad &\forall k \in \KKK \label{eq:UCB.1}\\
&\sum_{S\in \SSS}y(S) = 1, \quad y(S)\geq   0 &\quad & \forall S\in \SSS\label{eq:UCB.3}
\end{alignat}
\end{subequations}
for an optimal solution $\hat{y}_t$. The parameters in UCB-LP$(v_t, n_{t-1}, \omega)$ are defined as follows.
\begin{equation}\label{eq:epsilon}
\varepsilon(n) = \frac{(\sqrt{N} + 1)\Psi}{\sqrt{n}},
\end{equation}
\begin{equation}\label{eq:nt}
n_{t-1}(i) = \sum^{t-1}_{s = 1}\mathsf{1}(i\in S_t),
\end{equation}
\begin{equation}\label{eq:Psi}
\Psi = R(1 + B R)^2\sqrt{6\log{\frac{2NT(K+1)}{\delta}}},
\end{equation}
\begin{equation}\label{eq:omega}
\omega = \frac{11\Psi N}{\min_{k\in \KKK}c(k)}\sqrt{\frac{B}{T}\log\frac{4(K+1)}{\delta}}.
\end{equation}

By the assumption of $\omega < 1$ in Theorem \ref{thm:regret}, the right hand sides of the constraints (\ref{eq:UCB.1}) are positive. The linear program UCB-LP$(v_t, n_{t-1}, \omega)$ is always feasible, since $y(\emptyset) = 1$, $y(S) = 0$ for all $S\in\SSS\setminus \{\emptyset\}$ is always a feasible solution. Different from LP$(\hat{v})$ (which is used in {\sc Online}$(\tau)$), it is not known if UCB-LP$(v_t, n_{t-1}, \omega)$ can be efficiently solved (at least empirically) by the Column Generation algorithm or any other algorithm or heuristic.

The incorporation of confidence bounds into UCB-LP$(v_t, n_{t-1}, \omega)$ is inspired by \cite{AgrawalD14} as well as the primal-dual algorithm in \cite{BadanidiyuruKS13}. However, our design of the confidence bounds and the analysis are substantially different. As remarked in the design of {\sc Online}$(\tau)$, We cannot afford to learn all the choice probabilities $\{\varphi(i, S|v^*)\}_{i\in \NNN, S\in \SSS}$ individually, which would be the case if we just directly apply \cite{AgrawalD14}\cite{BadanidiyuruKS13}. Instead, we need to first provide a confidence bound on $v^*$, and then translate it to corresponding confidence bounds for the choice probabilities. The curse of dimensionality in learning is thus avoided. Different from {\sc Online}$(\tau)$, the confidence bounds for the UCB policy is adaptively defined every period. 


In the following, we outline the proof of Theorem \ref{thm:regret} in Appendix \ref{app:pfregretoutline}, and then prove the auxiliary Lemmas and Theorem in Appendices F - H.

\section{Proving the $\tilde{O}(\sqrt{T})$ Regret}\label{app:pfregretoutline}
\begin{procedure}[t]
\caption{Generation of $\{\tilde{S}_t, \tilde{I}_t\}^T_{t=1}$}\label{alg:tilde}
\begin{algorithmic}[1]
\For{$t = 1, \cdots, N$}
\State Define $\tilde{S}_i = S_i$, where $S_i$ are the fixed assortments defined in Line \ref{alg:UCB_line1} in Alg \ref{alg:UCB}.
\State Sample $\tilde{I}_t \sim \tilde{S}_t$.
\EndFor 
\For{$t = N+1, \cdots, T$}
\State Compute the MLE $\tilde{v}_t$ in (\ref{eq:-veloglikelihood}), based on $\{(\tilde{S}_s, \tilde{I}_s)\}^{t-1}_{s = 1}$.
\State Solve UCB-LP$(\tilde{v}_t, \tilde{n}_{t-1}, \omega)$ for an optimal $\tilde{y}_t$.
\State Select the sample assortment $\tilde{S}_t$ with prob. $\tilde{y}_t(\tilde{S}_t)$.\label{alg:tilde_Ut}
\State Sample $\tilde{I}_t\sim \tilde{S}_t$
\EndFor
\end{algorithmic}
\end{procedure}
First, we note that one particular challenge in analyzing the UCB policy is that it {\sc Abort}s at the random period $\tau$ when $C(k) = 0$. This makes the analysis of total revenue earned difficult. This is similar to the difficulty in analyzing {\sc Online}$(\tau)$.

Thus, to facilitate the analysis, we consider the following sales process $(\tilde{S}_t, \tilde{I}_t)^T_{t=1}$ generated by Procedure \ref{alg:tilde}. In Procedure \ref{alg:tilde}, the notation $\tilde{I}_t \sim \tilde{S}_t$ denotes sampling a product $\tilde{I}_t$ from $\tilde{S}_t\cup\{0\}$ with the underlying choice probability $\varphi(\tilde{I}_t, \tilde{S}_t|v^*)$. We define $\tilde{n}_{t-1}(i) = \sum^{t-1}_{s=1}\mathsf{1}(i\in \tilde{S}_s)$, similar to the definition of $n_{t-1}(i)$ in (\ref{eq:nt}). We emphasize that $(\tilde{S}_t, \tilde{I}_t)^T_{t=1}$ is only used for the analysis; the online algorithm does not need to know how to generate such a process. This is similar to the use of Procedure \ref{alg:tildesimple} for analyzing {\sc Online}$(\tau)$.

Note that $(\tilde{S}_t, \tilde{I}_t)^T_{t=1}$ is closely related to the sale process $(S_t, I_t)^T_{t=1}$ generated by Algorithm \ref{alg:UCB}. Let $t_{\text{stop}}$ be the period when Algorithm \ref{alg:UCB} signals {\sc Abort}; define $t_{\text{stop}} = T$ if no {\sc Abort} is signaled. When Algorithm \ref{alg:UCB} does not signal {\sc Abort}, the processes $(\tilde{S}_t, \tilde{I}_t, \tilde{v}_t, \tilde{n}_t )^T_{t=1}$ and $(S_t, I_t, v_t, n_t)^T_{t=1}$ are identically distributed. However, if an {\sc Abort} is signaled at period $t_{\text{stop}}$, then $(\tilde{S}_t, \tilde{I}_t, \tilde{v}_t, \tilde{n}_t)^{t_{\text{stop}}}_{t=1}$ and $(S_t, I_t, v_t, n_t)^{t_{\text{stop}}}_{t=1}$ are still identically distributed, but $S_t = \emptyset = I_t$ for $t \geq t_{\text{stop}} +1$, which is in general distributed differently from $(\tilde{S}_t, \tilde{I}_t)^T_{t=t_{\text{stop}} + 1}$. Moreover, our UCB policy satisfies the resource constraints with probability 1, i.e. $\sum^T_{t=1}a(I_t, k)\leq T c(k)$ with certainty; but $\sum^T_{t=1}a(\tilde{I}_t, k) > T c(k)$ violate the constraints with positive (despite being exponentially small) probability. 


Now, we have for any target regret bound {\sc Bound} the following inequality:
\begin{align}
&\PPP\left[\text{Regret}\leq  \text{{\sc Bound}}\right] \nonumber\\
= &\PPP\left[T\text{{\sc Opt}}(\text{LP}(v^*)) - \sum^{t_{\text{stop}}}_{t=1} r(I_t)\leq \text{{\sc Bound}}\right]\nonumber\\
\geq &\PPP\left[T\text{{\sc Opt}}(\text{LP}(v^*)) - \sum^T_{t=1} r(I_t)\leq \text{{\sc Bound}}, \text{no {\sc Abort}}\right] \nonumber\\
= &\PPP\left[\left\{T\text{{\sc Opt}}(\text{LP}(v^*)) - \sum^T_{t=1} r(\tilde{I}_t) \leq \text{{\sc Bound}}\right\}\right. \nonumber\\
&\quad \left. \cap \left\{\sum^T_{t=1}a(\tilde{I}_t, k)\leq Tc(k)\text{ for all $k$.}\right\}\right]. \nonumber
\end{align}
To prove Theorem \ref{thm:regret}, it suffices to prove the following two Lemmas:
\begin{lemma}\label{lemma:revenue}
We have 
\begin{align*}
&\PPP\left[T\text{{\sc Opt}}(\text{LP}(v^*)) - \sum^T_{t=1} r(\tilde{I}_t) \right. \nonumber\\
&\quad \qquad \left. = \tilde{O}\left(\sqrt{T} R^3 B^{5/2} N \right)\right]\geq 1-\frac{\delta}{K+1}.
\end{align*}
\end{lemma}
\begin{lemma}\label{lemma:constraints}
We have
\begin{align*}
&\PPP\left[\sum^T_{t=1} a(\tilde{I}_t, k) \leq Tc(k)\text{ for all $k\in \KKK$}\right] \geq 1-\frac{K\delta}{K+1}.
\end{align*}
\end{lemma}

The remaining exposition focuses on proving Lemmas \ref{lemma:revenue}, \ref{lemma:constraints}. To accomplish these tasks, we first prove the following instrumental Theorem, sheds light on the choice of parameter in $\text{UCB-LP}(v_t, n_{t-1}, \omega)$.
\begin{theorem}\label{thm:crucial}
Let $E_t$ denote the event that the inequality  
\begin{align}\label{eq:probconfbd}
& B(S |\tilde{v}_{t}) - \sum_{i\in S} \varepsilon(\tilde{n}_{t-1}(i))\leq B(S |v ^*)  \nonumber\\
&\quad \leq B( S |\tilde{v}_{t}) + \sum_{i\in S} \varepsilon(\tilde{n}_{t-1}(i))
\end{align}
holds for all $S\in \SSS$, $b\in [0, 1]^N$. (We use the notation $B( S | v ) = \sum_{i\in S}b(i)\varphi(i, S|v)$.) Then $E_t$ holds with probability at least $1-\delta/(2(K+1)T)$.
\end{theorem}
The proof of Theorem \ref{thm:crucial} is first outlined in Appendix \ref{app:crucial}, and the main Theorem in Appendix \ref{app:crucial} is proved in Appendix \ref{app:pfthmconfbd}. 

Finally, we prove Lemmas E.1, E.2 by using Theorem \ref{thm:crucial} in Appendices F, G. 

\subsection{Proving Theorem \ref{thm:crucial}}\label{app:crucial}
We prove Theorem \ref{thm:crucial} by establishing a confidence bound for estimating $v^*$, using correlated samples $\{\tilde{S}_s, \tilde{I}_s\}^{t-1}_{s = 1}$ generated by Algorithm \ref{alg:tilde}. Note that $\tilde{S}_t\in\sigma(\{\tilde{S}_s, \tilde{I}_s\}^{t-1}_{s = 1}\cup \tilde{U}_t)$, where $\tilde{U}_t$ is the randomness used to generate $\tilde{S}_t$ in Line \ref{alg:tilde_Ut}
\begin{theorem}\label{thm:confbd}
Consider the sales process $\{\tilde{S}_s, \tilde{I}_s\}^{t-1}_{s=1}$ generated by Algorithm \ref{alg:tilde}, where $t\geq N+1$. The following inequality
\begin{equation}\label{eq:confbd}
\sum^N_{i =1}\left(\sqrt{\tilde{n}_{t-1}(i)}\log \left|\frac{\tilde{v}_t(i)}{v^*(i)}\right| - \Psi\right)^2\leq N \Psi^2
\end{equation}
holds with probability at least $1-\delta/2T(K+1)$. 
\end{theorem}
The proof of Theorem \ref{thm:confbd} is postponed to Appendix \ref{app:pfthmconfbd}. 
The proof is similar to the proof of Lemma \ref{lemma:goodevent}, but the analysis in the proof of Theorem \ref{thm:confbd} is significantly more involved, since we need to disentangle the dependence across different products for the estimation of $v^*$.  

Similar to the proof of Lemma \ref{lemma:goodevent}, we consider the following change in variables $v(i) = e^{\theta(i)}$, and the function $L_t(\theta) = L_t((\theta(1), \ldots, \theta(N))) = \mathcal{L}_t((e^{\theta(1)}, \ldots, e^{\theta(N)})))$. The constant $\Psi$ is an artifact of the strong convexity of $L_t$ in $\theta$. A crucial part of the proof involves demonstrating the concentration property of $\nabla L_t(\theta^*)$, the gradient of $L_t$ at $\theta^* = (\theta^*(i))_{i\in \NNN} = (\log v^*(i))_{i\in \NNN}$. However, the classical Azuma-Hoeffding or Chernoof inequality is not directly applicable, since the frequency $\tilde{n}_{t-1}(i)$ is a random variable that correlates with $\{(\tilde{S}_s, \tilde{I}_s)\}^{t-1}_{s=1}$. This is in contrast to the analysis in the proof of Lemma \ref{lemma:goodevent}, where the number of observations on product $i$ is fixed to be $\tau/N$. Thus, we employ the following concentration inequality, which is commonly used in the multi-armed bandit literature:

\begin{lemma}[\cite{AbbasiPS11},\cite{BubeckMSS11}]\label{lemma:selfnorm}
Let $\{\FFF_t\}^\infty_{t = 1}$ be a filtration. Let $\rho(t)\in\{0, 1\}$ be a binary $\FFF_{t-1}$-measurable random variable, and let $\eta(t)$ be a $\FFF_{t}$-measurable random variable that is conditionally centered and $\FFF_{t-1}-$conditionally $L$-subGaussian, i.e. $\EEE[\eta(t)\mid \FFF_{t-1}] = 0$ a.s. and  $\EEE[e^{\lambda \eta(t)}\mid \FFF_{t-1}] \leq e^{(\lambda L)^2/2}$ for all $\lambda\in \RRR$. Then the confidence bound 
\begin{equation}
\left|\sum^\tau_{t = 1}\rho(t) \eta(t)\right|\leq L\sqrt{\left(1 + \sum^\tau_{t=1}\rho(t)\right)\left(1 + 2\log{\frac{\tau}{\delta}}\right)}
\end{equation}
holds with probability at least $1-\delta$.
\end{lemma}

The Lemma follows from either the application of Doob's Optional Sampling Theorem with Azuma-Hoeffding inequality (for example see the proof of Lemma 15 in \cite{BubeckMSS11}), or from the theory of self-normalizing processes (for example, see Lemma 6 in \cite{AbbasiPS11})

In particular, Theorem \ref{thm:confbd} implies that the confidence bound
\begin{equation}\label{eq:usefulbound}
\left|\log\frac{\tilde{v}_{t}(i)}{v^*(i)}\right| \leq  \varepsilon(\tilde{n}_{t-1}(i))
\end{equation}
holds for all product $i\in \NNN$ with probability at least $1-\delta/2T(K+1)$. 

Finally, combining (\ref{eq:usefulbound}) with Lemma \ref{lemma:lipschitz}, Theorem \ref{thm:crucial} is proved.

\section{Proof of Theorem \ref{thm:confbd}}\label{app:pfthmconfbd}
Recall that $\mathcal{L}_{t-1}(v)$ in the Theorem is the negative log-likelihood under the samples $\{\tilde{S}_s, \tilde{I}_s\}^{t-1}_{s=1}$ generated by Procedure \ref{alg:tilde}. (cf. (\ref{eq:-veloglikelihood})) Consider the the following change of variables and transformation on the likelihood function:
$$
\text{For all } i\in \NNN, \quad \theta(i) = \log v(i),
$$
\begin{equation*}
L_t(\theta) = L_t((\theta(1), \ldots, \theta(N))) = \mathcal{L}_t((e^{\theta(1)}, \ldots, e^{\theta(N)}))).
\end{equation*}
Also, we denote $\tilde{\theta}_t = (\log \tilde{v}_t(i))^{N}_{i=1}$, and $\theta^* = (\log v^*(i))^N_{i = 1}$.

By Taylor approximation, we know that there exists $\gamma\in[0,1]$ such that
\begin{align}\label{eq:taylor}
&L_{t-1}(\tilde{\theta}_t) = L_{t-1}(\theta^*) + \nabla L_{t-1}(\theta^*)^T(\tilde{\theta}_t - \theta^*)\nonumber\\
&\quad + \frac{1}{2}(\tilde{\theta}_t - \theta^*)^T H_{t-1}(\theta^* + \gamma(\tilde{\theta}_t - \theta^*))(\tilde{\theta}_t - \theta^*),
\end{align}
where 
$$\nabla L_{t-1}(\theta^*) = \left.\left(\frac{\partial L_{t-1}(\theta)}{\partial \theta(i)}\right)^N_{i=1}\right|_{\theta = \theta^*}$$ is the gradient at $\theta^*$, and
$$H_{t-1}(\theta^* + \gamma(\tilde{\theta}_t - \theta^*)) = \left.\left(\frac{\partial^2 L_{t-1}(\theta)}{\partial \theta(i)\partial\theta(j)}\right)_{1\leq i, j\leq N}\right|_{\theta = \theta^* + \gamma(\tilde{\theta}_t - \theta^*)}$$ is the Hessian matrix.

Now, we know that $L_{t-1}(\theta^*)\geq L_{t-1}(\tilde{\theta}_t)$, since $\tilde{v}_t$ minimizes $\mathcal{L}_t$. This  yields:
\begin{align}\label{eq:strongertaylor}
&\nabla L_{t-1}(\theta^*)^T(\tilde{\theta}_t - \theta^*) \nonumber\\
&\quad + \frac{1}{2}(\tilde{\theta}_t - \theta^*)^T H_{t-1}(\theta^* + s(\tilde{\theta}_t - \theta^*))(\tilde{\theta}_t - \theta^*) \leq 0.
\end{align}
Now, we claim the following two inequalities:
\begin{enumerate}
\item With probability at least $1-\delta/(2T(K+1))$, we have
\begin{equation}\label{eq:conc}
\left|\frac{\partial L_t}{\partial \theta(i)}\right|_{\theta = \theta^*} \leq \sqrt{6 \tilde{n}_{t-1}(i)\log\frac{2T(K+1)N}{\delta}}
\end{equation}
for all $i\in \NNN$.
\item We have 
\begin{align}\label{eq:psd}
H_{t-1}(\theta)&\succeq \frac{1}{R(1 + B R)^2}\times \nonumber\\
&\left( \begin{array}{ccccc}
n_{t-1}(1) & 0 & 0 & 0 & 0 \\
0 & n_{t-1}(2) & 0 & 0 & 0 \\
0 & 0 & \ddots &0 & 0 \\
0 &  0 & 0 & \ddots & 0 \\
0 & 0 & 0 & 0 & n_{t-1}(N) \\
\end{array} \right)
\end{align}
for all $\theta\in [-\log R, \log R]^N$. The notation $A\succeq B$ means that $A-B$ is positive semi-definite. 
\end{enumerate}
If (\ref{eq:conc}), (\ref{eq:psd}) hold, then we have the following from (\ref{eq:strongertaylor})
\begin{align*}
&\frac{1}{R(1 + B R)^2}\sum^N_{i=1}\left(\sqrt{\tilde{n}_{t-1}(i)}(\tilde{\theta}_t(i) - \theta^*(i))\right)^2 \nonumber\\ 
&\quad - \sqrt{6\log{\frac{2(K+1)TN}{\delta}}}\sum^N_{i=1} \sqrt{\tilde{n}_{t-1}(i)}\left|\tilde{\theta}_t(i) - \theta^*(i)\right| \leq 0.
\end{align*}
This leads to 
\begin{align*}
&\sum^N_{i=1}\left(\sqrt{\tilde{n}_{t-1}(i)}(\tilde{\theta}_t(i) - \theta^*(i))\right)^2 \nonumber\\
&\quad - 2\Psi\sum^N_{i=1} \sqrt{\tilde{n}_{t-1}(i)}\left|\tilde{\theta}_t(i) - \theta^*(i)\right| + \sum^{N}_{i=1}\Psi^2\leq N \Psi^2, 
\end{align*}
where we recall that $\Psi = R(1 + B R)^2\sqrt{6\log{\frac{2N(K+1)T}{\delta}}}$. This is what we are required to prove. 

To complete the proof, we prove (\ref{eq:conc}), (\ref{eq:psd}). $\textbf{Proving (\ref{eq:conc})}.$ The partial derivative has the following expression:
\begin{align}
\left.\frac{\partial L_t}{\partial \theta(i)}\right|_{\theta=\theta^*} &= \sum_{\substack{s\in \{1, \ldots, t-1\}: \\ \tilde{S}_s \ni i}}\varphi(i, \tilde{S}_s | v^*) - \mathsf{1}(\tilde{I}_s = i) \nonumber\\
&=\sum^{t-1}_{s=1} \rho_s(i)\left(\varphi(i, \tilde{S}_s | v^*) - \mathsf{1}(\tilde{I}_s = i)\right),\nonumber
\end{align}
where $\rho_s(i) = \mathsf{1}(\tilde{S}_s \ni i)$ is the indicator random variable of product $i$ being in the assortment $\tilde{S}_s$ in the $s^\text{th}$ period. Now, define the filtration $\FFF_{s-1} = \sigma(\{(\tilde{S}_\tau, \tilde{I}_\tau)\}^{s-1}_{\tau = 1}\cup \{\tilde{S}_s\})$, the $\sigma$-algebra generated by $\{(\tilde{S}_\tau, \tilde{I}_\tau)\}^{s-1}_{\tau = 1}\cup \{\tilde{S}_s\}$. Then the indicator $\rho_s(i)$ and the probability $\varphi(i, \tilde{S}_s | v^*)$ are $\FFF_{s-1}$-measurable, and the purchased product $\tilde{I}_s$ at period $s$ is $\FFF_s$-measurable. Now, we have $\EEE[\mathsf{1}(\tilde{I}_s = i)| \FFF_{s-1}] = \varphi(i, \tilde{S}_s | v^*)$, and clearly $\varphi(i, \tilde{S}_s | v^*) - \mathsf{1}(\tilde{I}_s = i)$ is 1-subGaussian. Thus, by applying Lemma \ref{lemma:selfnorm}, we have
\begin{align}
\left|\frac{\partial L_t}{\partial \theta(i)}\right|_{\theta = \theta^*} &= \left|\sum^{t-1}_{s=1} \rho_s(i)\left(\varphi(i, \tilde{S}_s | v^*) - \mathsf{1}(\tilde{I}_s = i)\right)\right| \nonumber\\
&\leq \sqrt{\left(1 + \tilde{n}_{t-1}(i)\right)\left(1 + 2\log\frac{2TN(K+1)}{\delta}\right)}\nonumber\\
&\leq \sqrt{6 \tilde{n}_{t-1}(i)\log\frac{2TN(K+1)}{\delta}} \label{eq:sqrt}
\end{align}
for all $i\in \NNN$ with probability at least $1-\delta/(2T(K+1))$, and the inequality (\ref{eq:sqrt}) is by the assumption that $\tilde{n}_{t-1}(i)\geq 1$ for all $i\in \NNN$.

$\textbf{Proving (\ref{eq:psd})}.$ First we express the second derivatives for $L_{t-1}$ for any $\theta\in \RRR^N$. For $i\neq j$,
\begin{equation}\label{eq:deri1}
\frac{\partial^2 L_{t-1}}{\partial \theta(i)\partial\theta(j)} = -\sum_{\substack{s\in \{1, \cdots, t-1\}: \\ \tilde{S}_s \ni i,j }}\frac{e^{\theta(i)}e^{\theta(j)}}{\left(1 + \sum_{\ell\in \tilde{S}_t }e^{\theta(\ell)}\right)^2}.
\end{equation}
\begin{equation}\label{eq:deri2}
\frac{\partial^2 L_{t-1}}{\partial \theta(i)^2} = \sum_{\substack{s\in \{1, \cdots, t-1\}: \\ \tilde{S}_s \ni i}}\frac{e^{\theta(i)} + \sum_{\ell\in \tilde{S}_t\setminus i}e^{\theta(i)}e^{\theta(\ell)}}{\left(1 + \sum_{\ell\in \tilde{S}_t }e^{\theta(\ell)}\right)^2}.
\end{equation}
Now, we focus on the Hessian matrix $h_s(\theta)$ for the $s^\text{th}$ period sample $(\tilde{S}_s, \tilde{I}_s)$. (We have $H_{t-1}(\theta) = \sum^{t-1}_{s=1} h_s(\theta)$.) By (\ref{eq:deri1}), (\ref{eq:deri2}), the Hessian matrix $h_s(\theta)$ can be 
expressed as follows:
\begin{align*}
&h_s(\theta) =  \frac{1}{(1 + \sum_{i\in \tilde{S}_s} e^{\theta(i)})^2} \times\nonumber\\
& \left( \begin{array}{ccccc}
e^{\theta(1)}\mathsf{1}(1\in \tilde{S}_s) & 0 & 0 & 0 & 0 \\
0 & e^{\theta(2)}\mathsf{1}(2\in \tilde{S}_s) & 0 & 0 & 0 \\
0 & 0 & \ddots &0 & 0 \\
0 &  0 & 0 & \ddots & 0 \\
0 & 0 & 0 & 0 & e^{\theta(N)}\mathsf{1}(N\in \tilde{S}_s) \\
\end{array} \right) \\
& \qquad + \sum_{\substack{1\leq i < j \leq N: \\ i, j\in \tilde{S}_s}}e^{\theta(i) + \theta(j)}u_{i,j}^T u_{i, j},
\end{align*}
where the vector $u_{i, j} = e_i - e_j$, and $e_i$ is the $i^\text{th}$ standard basis vector. Now, each term in the second summation is positive semi-definite. Applying the bound $v = e^{\theta(i)}\in [-R, R]$ for all $i\in \NNN$, and the model assumption that $|S|\leq B$ for all $S\in \SSS$, we have 
\begin{align}\label{eq:individualhessian}
& h_s(\theta) \succeq  \frac{1}{R(1 + B R)^2}\times \nonumber\\
&\left( \begin{array}{ccccc}
\mathsf{1}(1\in \tilde{S}_s) & 0 & 0 & 0 & 0 \\
0 & \mathsf{1}(2\in \tilde{S}_s) & 0 & 0 & 0 \\
0 & 0 & \ddots &0 & 0 \\
0 &  0 & 0 & \ddots & 0 \\
0 & 0 & 0 & 0 & \mathsf{1}(N\in \tilde{S}_s) \\
\end{array} \right),
\end{align}
and summing the inequality (\ref{eq:individualhessian}) over $1\leq s\leq t-1$ yields (\ref{eq:psd}). This concludes the proof of Theorem \ref{thm:confbd}.

\section{Proofs of Lemmas \ref{lemma:revenue}, \ref{lemma:constraints}}\label{app:remainingpfs}
\subsection{Proof of Lemma \ref{lemma:revenue}}\label{app:pflemmarevenue}
To upper bound the regret, we first have the following:
\begin{align}
&(1-\omega)(T-N)\text{{\sc Opt}}(\text{LP}(v^*)) - \sum^T_{t=N+1} r(\tilde{I}_t) \nonumber\\
\leq& \sum^{T}_{t = N+1}\sum_{S\in\SSS}\left(\sum_{i\in S} r(i) \varphi(i, S|\tilde{v}_t) + \varepsilon(\tilde{n}_{t-1}(i))\right)\tilde{y}_t(S) \nonumber\\
&\quad -\sum^T_{t=N+1}r(\tilde{I}_t)\label{eq:byclaim}\\
=& \sum^{T}_{t = N+1}\sum_{S\in\SSS}\left(\sum_{i\in S} r(i) \varphi(i, S|\tilde{v}_t) + \varepsilon(\tilde{n}_{t-1}(i))\right)\tilde{y}_t(S)\nonumber\\
&\underbrace{\qquad\qquad \quad - \sum^{T}_{t =N+ 1}\sum_{i\in \tilde{S}_t} r(i) \varphi(i, \tilde{S}_t|\tilde{v}_t) + \varepsilon(\tilde{n}_{t-1}(i))}_{(\diamondsuit_0)}\nonumber\\
\quad & + \sum^{T}_{t = N+1}\sum_{i\in \tilde{S}_t} r(i) \varphi(i, \tilde{S}_t|\tilde{v}_t) + \varepsilon(\tilde{n}_{t-1}(i))\nonumber\\
&\quad\underbrace{\qquad\qquad \quad  - \sum^{T}_{t = N+1}\sum_{i\in \tilde{S}_t} r(i) \varphi(i, \tilde{S}_t|v^*)}_{(\clubsuit_0)} \nonumber\\
\quad & \quad +\underbrace{\sum^{T}_{t = N+1}\sum_{i\in \tilde{S}_t} r(i) \varphi(i, \tilde{S}_t|\theta^*) - \sum^T_{t=N+1}r(\tilde{I}_t)}_{(\heartsuit_0)}.\nonumber 
\end{align}
The inequality (\ref{eq:byclaim}) is by the following Claim:
\begin{claim}\label{claim:upperbound}
Conditional on the event $E_t$ (recall $E_t$ from Theorem \ref{thm:crucial}), we have 
\begin{equation}
\text{{\sc Opt}}(\text{UCB-LP}(\tilde{v}_t, \tilde{n}_{t-1}, \omega)) \geq (1-\omega) \text{{\sc Opt}}(\text{LP}(v^*))
\end{equation}
\end{claim}
The proof of Claim \ref{claim:upperbound} is given in Appendix \ref{app:pfclaimupperbound}. It is similar to the proof of Lemma \ref{lemma:goodupperbound}. The claim shows that our UCB policy can indeed be seen as an optimism-in-face-of-uncertainty algorithm.

To bound $(\heartsuit_0)$: By our model assumption, $r(i)\in [0, 1]$ for all $i\in \NNN$. Observe that the $t^\text{th}$ summand $Rev_t =  \sum_{i\in \tilde{S}_t} r(i) \varphi(i, \tilde{S}_t|v^*) - r(\tilde{I}_t)\in[-1, 1]$ is a martingale difference with respect to the filtration $\FFF_t = \sigma(\{(\tilde{S}_s, \tilde{I}_s)\}^t_{s = 1}\cup \{\tilde{S}_{t+1}\})$, in the sense that $Rev_t$ is $\FFF_t$ measurable, and $\EEE[Rev_t | \FFF_{t-1}] = 0$. By applying Azuma-Hoeffding inequality, we have
\begin{equation}
(\heartsuit_0) \leq \sqrt{2T\log{\frac{4(K+1)}{\delta}}}
\end{equation}
with probability at least $1 - \delta/4(K+1)$.

To bound $(\clubsuit_0)$: We have the following bound: 
\begin{align}
(\clubsuit_0) &\leq 2\sum^T_{t = N+1}\sum_{i\in \tilde{S}_t}\varepsilon(\tilde{n}_{t - 1}(i))\quad \text{w. p. $\geq 1 - \frac{\delta}{2(K+1)}$}\label{eq:confbd0}\\
&\leq 2\Psi(\sqrt{N} + 1) \sum^{N}_{i = 1}\sum^{\tilde{n}_T(i)}_{n = 1}\frac{1}{\sqrt{n}}\label{eq:increment}\\
&\leq 4\Psi(\sqrt{N} + 1) \sum^{N}_{i = 1}\sqrt{\tilde{n}_T(i)}\nonumber\\
&\leq 4\Psi(\sqrt{N} + 1) \sqrt{N \sum^{N}_{i = 1}\tilde{n}_T(i)}\label{eq:jensen} \\
&\leq 4\Psi(\sqrt{N} + 1) \sqrt{N BT} < 8\sqrt{T} \Psi \sqrt{B} N .\label{eq:bdass}
\end{align}
The inequality (\ref{eq:confbd0}) holds with probability at least $1- \delta/2(K+1)$, by Theorem \ref{thm:crucial} and a union bound over the periods. All inequalities apart from (\ref{eq:confbd0}) hold with probability 1. The inequality (\ref{eq:increment}) is based on the following observation. Fix a particular product $i$, and let $\tilde{S}_{t_1}, \cdots, \tilde{S}_{t_m}$ be the assortments that includes $i$ from period $N+1$ to period $T$, where $N+1 \leq t_1< t_2 < \cdots < t_m$. The summand $\varepsilon(\tilde{n}_{t-1}(i))$ appears in (\ref{eq:confbd0}) at each of the time indexes $t_j$, and it is clear that $\tilde{n}_{t_{j+1} - 1}(i) = \tilde{n}_{t_j - 1}(i) + 1$. The inequality (\ref{eq:jensen}) is by Jensen's Inequality. Finally, the inequality (\ref{eq:bdass}) is by the fact that at most $B$ products can be included in each of the $T$ assortments.
 
To bound $(\diamondsuit_0)$: By the definition of $\varepsilon$ in (\ref{eq:epsilon}), we have $\left|\sum_{i\in S} r(i) \varphi(i, S|\tilde{v}_t) + \varepsilon(\tilde{n}_{t-1}(i))\right| \leq 2B\sqrt{N}\Psi$ for all $i\in S\in\SSS$ and all $t$. \footnote{Better bound can be proved, but it does not affect the overall regret in the analysis.} Observe that the $t^\text{th}$ summand $rev_t =  \sum_{S\in\SSS}\left(\sum_{i\in S} r(i) \varphi(i, S|\tilde{v}_t) + \varepsilon(\tilde{n}_{t-1}(i))\right)\tilde{y}_t(S)- \sum_{i\in \tilde{S}_t} r(i) \varphi(i, \tilde{S}_t|\tilde{v}_t) + \varepsilon(\tilde{n}_{t-1}(i))$ is a martingale difference with respect to the filtration $\eee_t = \sigma(\{(\tilde{S}_s, \tilde{I}_s)\}^t_{s = 1})$. This is because $rev_t$ is $\eee_t$ measurable ($\tilde{n}_{t-1}(i), \tilde{y}_t$ are $\eee_t$ measurable), and $\EEE[rev_t | \eee_{t-1}] = 0$. By applying Azuma-Hoeffding inequality, we have
\begin{align*}
&(\diamondsuit_0) \leq 2\sqrt{N}B\Psi\sqrt{2T\log{\frac{4(K+1)}{\delta}}}\nonumber\\
\leq & 2N\Psi\sqrt{2BT\log\frac{4(K+1)}{\delta}}
\end{align*}
with probability at least $1 - \delta/4(K+1)$.

So the regret in the revenue is at most 
\begin{align}
&N +  \omega  T\text{{\sc Opt}}(\text{LP}(v^*))+ (\diamondsuit_0) + (\clubsuit_0) + (\heartsuit_0) \nonumber\\
< &  11\Psi N\sqrt{BT\log{\frac{4(K+1)}{\delta}}}\left(1 + \frac{1}{\min_{k\in \KKK}c(k)}\right) \label{eq:preciseregret}\\
= &O\left(\sqrt{T} R^3  B^{5/2}N\log{\frac{NTK}{\delta}}\right)\nonumber .
\end{align} 
with probability at least $1-\delta/(K+1)$.

\section{Proof of Claim \ref{claim:upperbound}}\label{app:pfclaimupperbound}
Let $y^*$ be an optimal solution to LP$(v^*)$, and consider the solution $\bar{y} = (1-\omega)y^* + \omega \mathsf{1}_{\emptyset}$. That is $\bar{y}(S) = (1-\omega)y^*(S)$ for $S\in\SSS\setminus \{\emptyset\}$, and $\bar{y}(\emptyset) = (1-\omega)y^*(S) + \omega$. First, it is clear that $\bar{y}$ is feasible to UCB-LP$(\tilde{v}_t, \tilde{n}_{t-1}, \omega)$. Clearly, $\bar{y} \geq 0$, and $\sum_{S\in\SSS}\bar{y}(S) = (1-\omega)\sum_{S\in\SSS}y^*(S) + \omega = 1.$ Moreover, for each resource $k\in \KKK$, we have 
\begin{align}
&\sum_{S\in \SSS} \left(\sum_{i\in S}a(i, k)\varphi(i, S\mid \tilde{v}_t) - \varepsilon(\tilde{n}_{t-1}(i))\right)\bar{y}(S) \nonumber\\
=&(1 - \omega)\sum_{S\in \SSS} \left(\sum_{i\in S}a(i, k)\varphi(i, S\mid \tilde{v}_t) - \varepsilon(\tilde{n}_{t-1}(i))\right)y^*(S)\label{eq:nullproduct}\\
\leq & (1 - \omega)\sum_{S\in \SSS} \left(\sum_{i\in S}a(i, k)\varphi(i, S\mid \tilde{v}_t) - \varepsilon(\tilde{n}_{t-1}(i))\right)y^*(S)  \label{eq:defofLCB}\\
\leq &(1-\omega)c(k) \nonumber,
\end{align}
where inequality(\ref{eq:nullproduct}) is by the fact that $a(0, k)=  0$ for all $k\in \KKK$, inequality (\ref{eq:defofLCB}) is by the definition of event $E_t$. (Recall $E_t$ from Theorem \ref{thm:crucial})

Since $\tilde{y}_t$ is optimal for UCB-LP$(\tilde{v}_t, \tilde{n}_{t-1}, \omega)$, we have
\begin{align}
&\text{{\sc Opt}}(\text{UCB-LP}(\tilde{v}_t, \tilde{n}_{t-1}, \omega)) \nonumber\\
=& \sum_{S\in \SSS} \left(\sum_{i\in S}r(i) \varphi (i, S\mid \tilde{v}_t) + \varepsilon(\tilde{n}_{t-1}(i))\right)\tilde{y}_t(S) \nonumber\\
\geq &\sum_{S\in \SSS} \left(\sum_{i\in S}r(i) \varphi (i, S\mid \tilde{v}_t) + \varepsilon(\tilde{n}_{t-1}(i))\right)\bar{y}(S)\label{eq:ybarfeasible}\\
\geq & \sum_{S\in \SSS} \sum_{i\in S}r(i) \varphi(i, S\mid v^*)\bar{y}(S) \label{eq:defofUCB}\\
= & (1-\omega) \text{{\sc Opt}}(\text{LP}(v^*))\nonumber.
\end{align}
Inequality (\ref{eq:ybarfeasible}) is by the feasibility of $\bar{y}$ to UCB-LP$(\tilde{v}_t, \tilde{n}_{t-1}, \omega)$, and inequality (\ref{eq:defofUCB}) is by the definition of $E_t$. This proves the Theorem.

\subsection{Proof of Lemma \ref{lemma:constraints}}\label{app:pflemmaconstraints}

For each $k\in \KKK$, we have the following:
\begin{align}
& \sum^{T}_{N+1}a(\tilde{I}_t, k) \nonumber\\
=& \underbrace{\sum^T_{t=N+1}a(\tilde{I}_t, k) - \sum^{T}_{t = N+1}\sum_{i\in \tilde{S}_t} a(i, k) \varphi(i, \tilde{S}_t|v^*)}_{(\heartsuit_k)} \nonumber\\
& +\sum^{T}_{t = N+1}\sum_{i\in \tilde{S}_t} a(i, k) \varphi(i, \tilde{S}_t|v^*) \nonumber\\
& \quad \underbrace{\qquad - \sum^{T}_{t = N+1}\sum_{i\in \tilde{S}_t} a(i,k) \varphi(i, \tilde{S}_t|\tilde{v}_t) - \varepsilon(\tilde{n}_{t-1}(i))  }_{(\clubsuit_k)} \nonumber\\
\quad &  + \sum^{T}_{t = N+1}\sum_{i\in \tilde{S}_t} a(i,k) \varphi(i, \tilde{S}_t|\tilde{v}_t) - \varepsilon(\tilde{n}_{t-1}(i)) \nonumber\\
\quad & \quad \underbrace{ - \sum^{T}_{t = N+1}\sum_{S\in\SSS}\left(\sum_{i\in S} a(i,k) \varphi(i, S|\tilde{v}_t) - \varepsilon(\tilde{n}_{t-1}(i))\right)\tilde{y}_t(S)}_{(\diamondsuit_k)} \nonumber\\
& + \underbrace{\sum^{T}_{t = N+1}\sum_{S\in\SSS}\left(\sum_{i\in S} a(i,k) \varphi(i, S|\tilde{v}_t) - \varepsilon(\tilde{n}_{t-1}(i))\right)\tilde{y}_t(S)}_{(\spadesuit_k)} \nonumber 
\end{align}

To bound $(\heartsuit_k)$: By our model assumption, $a(i, k)\in \{0, 1\}$ for all $i\in \NNN, k\in \KKK$. Observe that the $t^\text{th}$ summand $A_t = a(\tilde{I}_t, k) - \sum_{i\in \tilde{S}_t} a(i, k) \varphi(i, \tilde{S}_t|v^*)\in[-1, 1]$, and the summands $\{A_t\}^T_{t=N+1}$ is a martingale difference sequence with respect to the filtration $\FFF_t = \sigma(\{(\tilde{S}_s, \tilde{I}_s)\}^t_{s = 1}\cup \{\tilde{S}_{t+1}\})$, in the sense that $A_t$ is $\FFF_t$ measurable, and $\EEE[A_t | \FFF_{t-1}] = 0$. By applying Azuma-Hoeffding inequality, we have
\begin{equation}
(\heartsuit_k) \leq \sqrt{2T\log{\frac{4(K+1)}{\delta}}}
\end{equation}
with probability at least $1 - \delta/(4(K+1))$.

To bound $(\clubsuit_k)$: We have
\begin{equation*}
(\clubsuit_k) \leq 2\sum^T_{t = N+1}\sum_{i\in \tilde{S}_t}\varepsilon(\tilde{n}_{t - 1}(i))< 8\Psi N \sqrt{BT},
\end{equation*}
with probability at least $1- \delta/(2(K+1))$, where the first inequality is by Theorem \ref{thm:confbd} and our model assumption that $a(i, k)\in \{0,1\}$ for all $i\in \NNN, k\in \KKK$, and the second inequality follows exactly the same reasoning as in $(\clubsuit_0)$.

To bound $(\diamondsuit_k)$: By the definition of $\varepsilon$ in (\ref{eq:epsilon}), we have $\left|\sum_{i\in S} a(i,k) \varphi(i, S|\tilde{v}_t) - \varepsilon(\tilde{n}_{t-1}(i))\right| \leq 2B\sqrt{N}\Psi$ for all $i\in S\in\SSS$ and all $t$. Observe that the $t^\text{th}$ summand $a_t = \sum_{i\in \tilde{S}_t} a(i,k) \varphi(i, \tilde{S}_t|\tilde{v}_t) - \varepsilon(\tilde{n}_{t-1}(i)) - \sum_{S\in\SSS}\left(\sum_{i\in S} a(i,k) \varphi(i, S|\tilde{v}_t) - \varepsilon(\tilde{n}_{t-1}(i))\right)\tilde{y}_t(S) $ is a martingale difference with respect to the filtration $\eee_t = \sigma(\{(\tilde{S}_s, \tilde{I}_s)\}^t_{s = 1})$. This is because $a_t$ is $\eee_t$-measurable ($\tilde{n}_{t-1}(i), \tilde{y}_t$ are $\eee_t$-measurable), and $\EEE[a_t | \eee_{t-1}] = 0$. By applying Azuma-Hoeffding inequality, we have
\begin{align*}
&(\diamondsuit_k) \leq 2\sqrt{N}B\Psi\sqrt{2T\log{\frac{4(K+1)}{\delta}}} \nonumber\\
\leq & 2N\Psi\sqrt{2BT\log\frac{4(K+1)}{\delta}}
\end{align*}
with probability at least $1 - \delta/(4(K+1))$.

Total amount $\sum^T_{t=1}a(\tilde{I}_t, k)$ of resource $k$ consumed from Period 1 to period $T$ is at most
\begin{align*}
&N + (\diamondsuit_k) + (\clubsuit_k) + (\heartsuit_k) + (\spadesuit_k)\\ 
< & 11\Phi N\sqrt{BT\log{\frac{4(K+1)}{\delta}}} + (\spadesuit_k) \\
< &11\Phi N\sqrt{BT\log{\frac{4(K+1)}{\delta}}} + T(1-\omega)c(k) \leq Tc(k).
\end{align*}
with probability at least $1- \delta/(K+1)$. That is $C(k) > 0$ for all $k\in \KKK$ and for all periods with probability at least $1- K\delta/(K+1)$. Thus the Lemma is proved.

\bibliographystyle{named}
\bibliography{assortmentref}

\end{document}